\documentclass[sn-aps, dvipsnames, table]{sn-jnl}

\usepackage{color}
\usepackage{xcolor}

\usepackage[figuresright]{rotating}
\usepackage{longtable}
\usepackage{diagbox}
\usepackage[justification=centering]{caption}
\usepackage{hyperref}
\usepackage{bbding}
\usepackage{graphicx}

\usepackage{indentfirst}
\usepackage{enumerate}
\usepackage{makecell}
\usepackage{multirow}
\usepackage{float}
\newcommand{\xgy}[1]{{\color{black} #1}}

\newcommand{\reviewer}[1]{{\color{black} #1}}

\jyear{2022}%

\theoremstyle{thmstyleone}%
\theoremstyle{thmstyletwo}%

\theoremstyle{thmstylethree}%

\begin{document}

\title[Deep Industrial Image Anomaly Detection: A Survey]{Deep Industrial Image Anomaly Detection: A Survey}


\author[1]{\fnm{Jiaqi} \sur{Liu}}
\author[1,2]{\fnm{Guoyang} \sur{Xie}}
\author[1]{\fnm{Jinbao} \sur{Wang}}

\author[1]{\fnm{Shangnian} \sur{Li}}
\author[3]{\fnm{Chengjie} \sur{Wang}}

\author[1]{\fnm{Feng} \sur{Zheng}}
\equalcont{Corresponding Authors.}
\author[2,4]{\fnm{Yaochu} \sur{Jin}}
\equalcont{Corresponding Authors.}





\affil[1]{\orgdiv{Research Institute of Trustworthy Autonomous Systems}, \orgname{Southern University of Science and Technology}, \orgaddress{\city{Shenzhen} \postcode{518055},  \country{China}}}

\affil[2]{\orgdiv{NICE Group}, \orgname{University of Surrey}, \orgaddress{\city{Guildford} \postcode{GU2 7YX},  \country{United Kingdom}}}

\affil[3]{\orgdiv{Youtu Lab}, \orgname{Tencent}, \orgaddress{\city{Shanghai}  \postcode{200233},  \country{China}}}

\affil[4]{\orgdiv{NICE Group}, \orgname{Bielefeld University}, \orgaddress{\city{Bielefeld}  \postcode{33619},  \country{Germany}}}



\abstract{

The recent rapid development of deep learning has laid a milestone in industrial Image Anomaly Detection (IAD). In this paper, we provide a comprehensive review of deep learning-based image anomaly detection techniques, from the perspectives of neural network architectures, levels of supervision, loss functions, metrics and datasets. \reviewer{In addition, we extract the promising setting from industrial manufacturing and review the current IAD approaches under our proposed setting.} Moreover, we highlight several opening challenges for image anomaly detection. The merits and downsides of representative network architectures under varying supervision are discussed. Finally, we summarize the research findings and point out future research directions. More resources are available at \href{https://github.com/M-3LAB/awesome-industrial-anomaly-detection}{https://github.com/M-3LAB/awesome-industrial-anomaly-detection}. 
}

\keywords{Image anomaly detection, Defect detection, Industrial manufacturing, Deep learning, Computer vision}



\maketitle

\section{Introduction}\label{sec1}
We review the recent advances of deep learning-based image anomaly detection since the rapid development of deep learning can bring the capabilities of image anomaly detection into the factory floor. In modern manufacturing, IAD is always performed at the end of the manufacturing process and tries to identify product defects. The price of a product is significantly affected by the defect's severity. In addition, if the flaw reaches a certain threshold, the product will be discarded. Historically, the majority of anomaly detection tasks are performed by humans, which suffers from the following many disadvantages:
\begin{itemize}
    \item It is impossible to avoid human fatigue, resulting in a false positive phenomenon (\textit{i.e.}, the ground truth is abnormal, while the human's judgment is normal).
    \item Long and intensive work on anomaly detection may cause health problems, such as visual impairment.
    \item Locating anomalies requires a significant number of employees, raising operational costs.
\end{itemize}
Thus, the goal of IAD algorithms is to reduce human labour and improve productivity and product quality. Before deep learning, the performance of IAD could not fulfil the demands of industrial manufacturing. Nowadays, the deep learning method has received good results, and most of these methods are more than 97\% accurate. Still, IAD has many problems when it comes to real-world use. To comprehensively explore the effectiveness and applicable scenarios of the current methods, more careful analysis of IAD we conduct in this survey is necessary and significant.  

\begin{table}[htbp]
\centering
\renewcommand{\arraystretch}{1.2}
\caption{Related surveys and ours for IAD.}
\resizebox{0.85\textwidth}{!}{
\begin{tabular}{l|cccc}
\hline
\rowcolor{NavyBlue!10} \textbf{Content }    & Czimmermann~\cite{czimmermann2020visual}  & Tao~\cite{tao2022deep} & Cui~\cite{cui2022survey} & \textbf{Ours} \\ \hline
IAD dataset            & -        & 9     & 7       & \textbf{20}   \\
IAD metric             & -        & 3     & 1       & \textbf{6}    \\
Neural network architecture            &   \textcolor{red}{\XSolidBrush}             & \textcolor{green}{\Checkmark}     & \textcolor{green}{\Checkmark}       & \textcolor{green}{\Checkmark}    \\
Levels of supervision & \textcolor{green}{\Checkmark}              & \textcolor{red}{\XSolidBrush}     & \textcolor{red}{\XSolidBrush}       & \textcolor{green}{\Checkmark}    \\
Industrial manufacturing setting         & \textcolor{red}{\XSolidBrush}               & \textcolor{red}{\XSolidBrush}     & \textcolor{red}{\XSolidBrush}       & \textcolor{green}{\Checkmark}   \\ \hline

\end{tabular}
}
\label{tab:survey_comparison}
\end{table}

Table~\ref{tab:survey_comparison} demonstrates clearly the merits of our survey in terms of dataset, metric, neural network architecture, levels of supervision and promising setting for industrial manufacturing. As a representative review that focuses more on traditional methods, Czimmermann et al.~\cite{czimmermann2020visual} have less discussion of deep learning methods, while our survey discusses deep learning in more depth. Firstly, our study uses twice as many IAD datasets as Tao~\textit{et al.}~\cite{tao2022deep}. Secondly, we analyze the performance of IAD using the most comprehensive image level and pixel level metrics. Nevertheless, Cui~\textit{et al.}~\cite{cui2022survey} and Tao~\textit{et al.}~\cite{tao2022deep} only employ image level metrics, neglecting the anomalies localization performance of IAD. Thirdly, our study develops a taxonomy based on the design of neural network architecture with varying degrees of supervision. Finally, to bridge the gap between academic research and real-world industry needs, we review the current IAD algorithms under industrial manufacturing settings.


As an emerging field, research on IAD must fully consider industrial manufacturing requirements. The following is a summary of the challenging issues that need to be investigated:
\begin{itemize}
    \item IAD dataset should be gathered from actual manufacturing lines, not labs. The public cannot access the real-world anomalous dataset due to privacy concerns. The majority of open-source IAD datasets generate anomalies from anomaly-free products. In other words, the abnormalities from open-source IAD datasets may not occur in actual production lines, which makes deploying IADs in industrial manufacturing very challenging. 
    \item It is challenging to enable the creation of a unified IAD model in the absence of multiple domain IAD datasets. Recently, You~\textit{et al.}~\cite{you2022unified} propose a unified IAD model for multiple class objects. However, they disregard the notion that commodities produced in the same plant should be of the same sort. For example, an automaker manufactures several types of workpieces but does not produce fruit. Current popular IAD datasets, like MVTec AD~\cite{bergmann2019mvtec} and MVTec LOCO~\cite{bergmann2022beyond}, consist of numerous classes but not multiple domains. To simulate a realistic manufacturing process, we must create a new IAD dataset collected from multiple domains. 
    \item It is urgent to set up a uniform assessment for the image-level and pixel level of IAD performance. The majority of IAD metrics shrink the anomalous mask (ground truth) into the size of feature map for evaluation, which inevitably reduces the precision of assessment. Moreover, we discover that certain IAD methods perform well on image AUROC but poorly on pixel AP, or vice versa. Therefore, it is essential to develop a uniform metric for assessment IAD performance at both image and pixel level.
    \item We should design a more efficient loss function that can leverage both the guidance of labelled data and the exploration of unlabelled data. In realistic manufacturing scenario, limited number of anomalous samples are available. However, most of unsupervised IAD methods outperform semi-supervised IAD methods. By observing the failure of semi-supervised IAD, we would call for more attention to the feature extraction and loss function, which can leverage both the guidance from labels efficiently and the exploration from the unlabeled data. Regarding the key problem mentioned above, improving feature extraction from abnormal samples and redesigning deviation loss function can fully use labelled anomalies and diverge the feature space of abnormal samples from those of normal samples. 
\end{itemize}

\reviewer{The paper categorizes various methods into several paradigms, and clearly analyzes the advantages and disadvantages of various paradigms. It allows the reader to understand the state-of-the-art quickly and provides a reliable guide for selecting the required algorithm for practical applications. More importantly, we have analyzed the disadvantages of different paradigms and the current main challenges. Subsequent researchers can quickly find directions to push the field forward.}

\subsection{Contributions}
The main contributions of this survey can be summarized as following:
\begin{itemize}
    \item We provide an in-depth review of image anomaly detection by considering the design of neural network architecture with varying degrees of supervision.
    \item \reviewer{It provides a comprehensive review of the current IAD algorithms in different settings to bridge the gap between the academic research and real-world industrial manufacturing.}
    \item It summarizes the main issues and potential challenges in IAD, which outlines the underlying research directions for future works.
\end{itemize}

\begin{figure}[t]
    \centering
    \includegraphics[width=\linewidth]{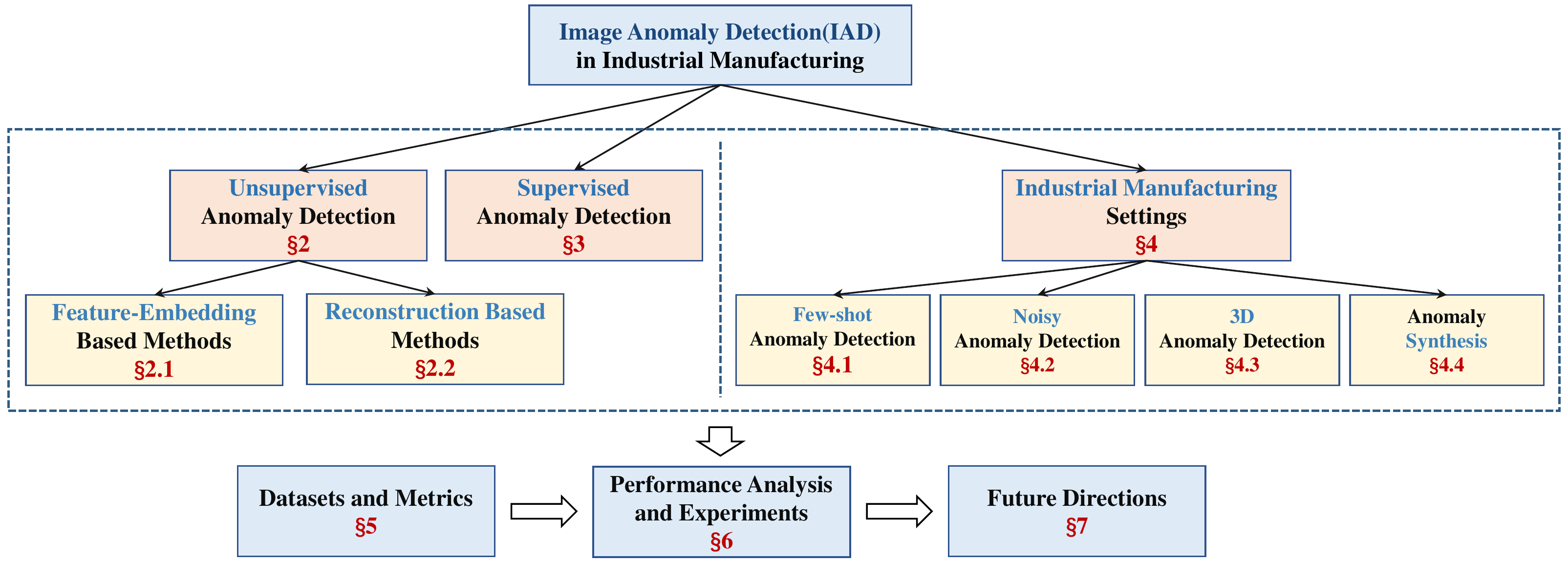}
    \caption{Framework of this survey.}
    \label{fig:framework_survey}
\end{figure}

The rest of this paper is organized as Figure~\ref{fig:framework_survey}. In Section~\ref{sec:unsupervised} and Section~\ref{sec:supervised_detection}, we review IAD on the basis of the neural network architecture with different levels of supervision. Next, we review the recent advances of IAD under our proposed setting from industrial manufacturing in Section~\ref{sec:manufacturing_setting}. We describe the popular dataset in Section~\ref{sec:data} and take a retrospective view of the metrics function in Section~\ref{sec:metrics}. Then, we provide an analysis of the performance of current IAD methods on various datasets in Section~\ref{sec:total_performance}. Finally, we provide future research directions for IAD in Section~\ref{sec:future_direction}. 




\section{Unsupervised Anomaly Detection}\label{sec:unsupervised}


The majority of current research focuses on unsupervised anomaly detection, based on the assumption that the collection of abnormal samples incurs massive human and financial costs. This indicates that only normal samples are included in the training set, whereas both abnormal and normal samples are included in the test set. Anomaly detection in industrial images is a subset of problems with out-of-distribution (OOD). Before the rise of deep learning, differential detection and filtering were frequently used to detect anomalies in industrial images. Following the release of the MVTec AD~\cite{bergmann2019mvtec}, methods for anomaly detection in industrial images can be divided into two categories: feature-embedding and reconstructed-based. Currently, more AD techniques are based on feature embedding.

%
\subsection{Feature Embedding based Methods}
\subsubsection{Teacher-Student Architecture}

 
 The performance of these methods is outstanding, but they depend on pre-trained models such as ResNet~\cite{he2016deep} VGG~\cite{simonyan2014very} and EfficientNet~\cite{tan2019efficientnet}.
The selection of the ideal teacher model is crucial. This type of instructional strategy is summarized in Table~\ref{tab:teacher_student_summary}. The structure of the network and the method of distillation are the primary distinctions between various techniques.

\begin{table}[ht]
\centering
\caption{A summary of teacher-student methods regarding loss function, pre-trained model, and highlights.}
\renewcommand{\arraystretch}{1.2}
\resizebox{\textwidth}{!}{
\begin{tabular}{p{2cm} p{3cm}<{\raggedright} p{2cm} p{10cm}}
\hline
 \rowcolor{NavyBlue!10} \textbf{Method}      & \textbf{Loss Function }   & \textbf{Pre-trained}       & \textbf{Highlights}       \\ \hline 
Uninformed Students~\cite{bergmann2020uninformed}
            & $L_2$, Compactness           & ResNet             & The paper designs a basic approach to anomaly detection problems using a teacher-student model.                           \\
MKD~\cite{salehi2021multiresolution}         & $L_2$      & VGG                & The paper uses multi-scale features and lighter networks for distillation.                              \\
STPM~\cite{Wang2021StudentTeacherFP}                        & $L_2$                  & ResNet             & The paper uses multi-scale features under different network layers for distillation.                    \\
STFPM~\cite{yamada2021reconstruction}                       & $L_2$                 & ResNet            & The paper adds another teacher-student pair to get different feature reconstruction results.            \\
RD4AD~\cite{deng2022anomaly}                       & Cosine Similarity               & ResNet             & The paper designs the teacher-student model of reverse distillation in a similar way to reconstruction. \\
IKD~\cite{cao2022informative}                         & Context Similarity           & ResNet             & The paper adds context similarity loss and adaptive hard sample mining module to prevent overfitting.                             \\
AST~\cite{rudolph2022asymmetric}            & $L_2$, Log-Likelihood & EfficientNet       & The paper uses a asymmetric  teacher-student network to make the representation of anomaly more different. \\ \hline
\end{tabular}  
}
\label{tab:teacher_student_summary}
\end{table}

\begin{figure}[ht]
\centering
    \includegraphics[width=0.75\linewidth]{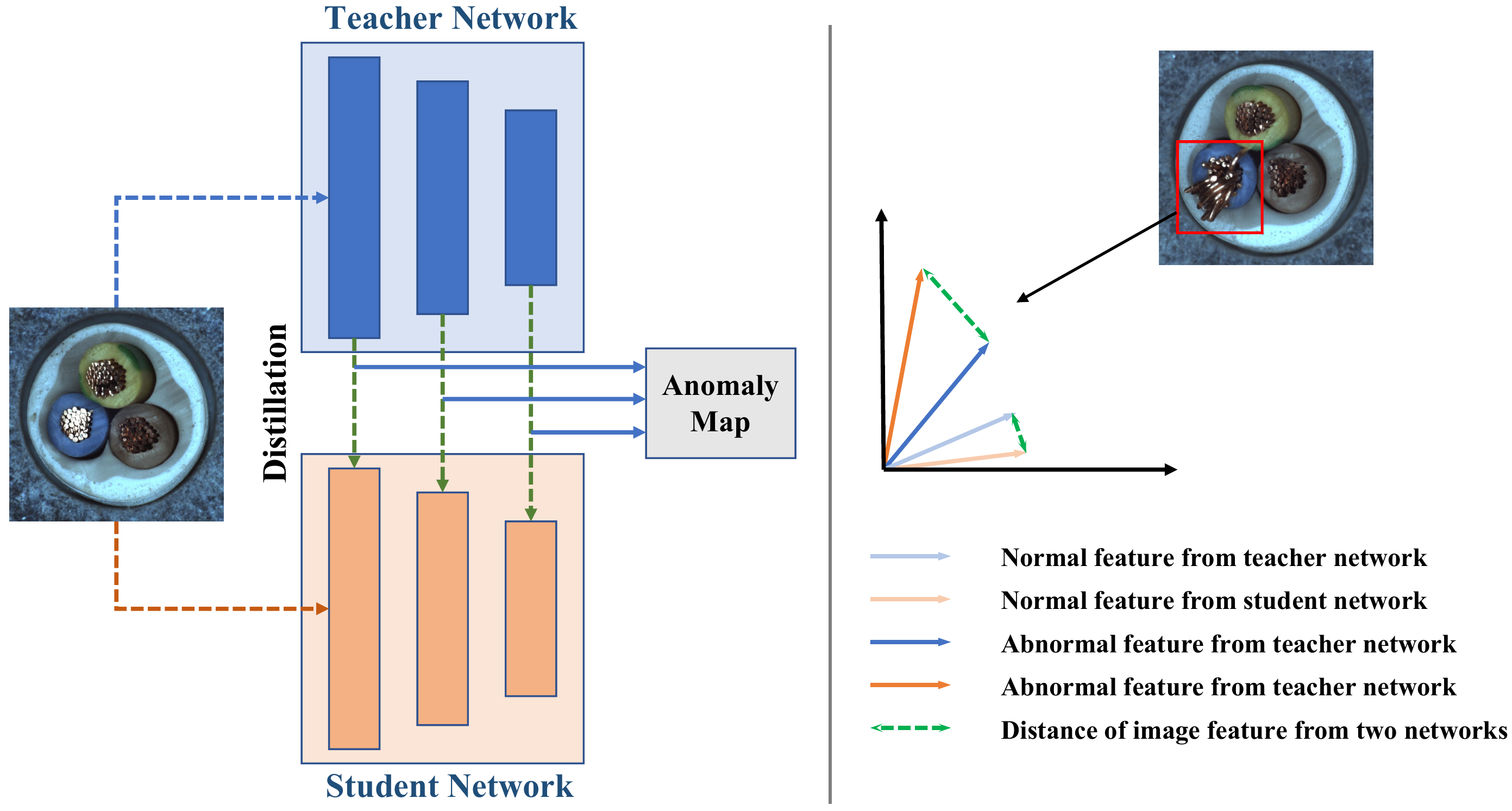}
    \caption{Architecture of teacher-student models.}
    \label{fig:teacher-student}
\end{figure}

The teacher-student network architecture depicted in Fig.~\ref{fig:teacher-student} is the most standard technique for detecting industrial image anomalies. This method typically selects a partial layer of a backbone network pre-trained on a large-scale dataset as a fixed-parameter teacher model. During training, the teacher model imparts to the student model the knowledge of extracting normal sample features. During inference, the characteristics of normal images extracted from the test set by the teacher network and the student network are comparable, whereas the characteristics of abnormal images extracted from the test set are quite distinct. By comparing the feature maps generated by the two networks, it is possible to generate anomaly score maps with the same size. Then, by enlarging the anomaly score map to the same proportion as the input image, we can obtain the anomaly scores of various input image locations. On the justification of this model, it is possible to determine whether the test image is abnormal.

Bergmann \textit{et al.}~\cite{bergmann2020uninformed} is the first to use teacher-student architecture for anomaly detection. The model is straightforward and effective, significantly outperforming other benchmark methods. While STPM~\cite{Wang2021StudentTeacherFP} and MKD~\cite{salehi2021multiresolution} both use multi-scale features under different network layers for distillation, they do so in different ways. In this instance, the normal sample features extracted by the student network are more similar to those extracted by the teacher network, whereas the abnormal sample features are more dissimilar.
In addition, MKD finds that the lighter student network structure performs better than the student network structure identical to that of the teacher network. Based on STPM, RSTPM~\cite{yamada2021reconstruction,yamada2022reconstructed} adds a pair of teacher-student networks. During reasoning, the new teacher network is placed behind the original teacher-student network and is responsible for recreating the features. When anomalous images are presented, the student network typically reconstructs normal features that can be distinguished from those of the teacher network.
RSTPM also includes a mechanism for transferring features from the teacher network to the student network in order to facilitate feature reconstruction. RD4AD~\cite{deng2022anomaly} and RSTPM share certain similarities in their learning. RSTPM employs two pairs of teacher-student networks for feature reconstruction, whereas RD4AD only employs one pair of teacher-student networks. RD4AD proposes a Multi-scale Feature Fusion (MFF) block and One-Class Bottleneck (OCB) to form an embedding, which is used to eliminate redundant features at multiple scales so that a single pair of teacher-student networks can perform feature reconstruction effectively. The abnormal image features extracted by the teacher-student network of RD4AD differ significantly during inference.
AST~\cite{rudolph2022asymmetric} concludes that the abnormal image features extracted by the teacher-student model with the same structure are significantly similar, so they propose an asymmetric teacher-student architecture to address this issue. AST also introduces a normalized flow to avoid this problem and prevent estimation bias caused by the inconsistency of the two network structures. Previous teacher-student architecture anomaly detection methods suffer from overfitting as a result of inconsistency between neural network capacity and knowledge amount. By incorporating the Context Similarity Loss (CSL) and Adaptive Hard Sample Mining (AHSM) modules, Informative Knowledge Distillation (IKD)~\cite{cao2022informative} hopes to reduce overfitting. CSL can assist the student network in comprehending the structure of a context-containing data manifold. The AHSM can concentrate on difficult samples containing a lot of information.

\subsubsection{One-Class Classification}

One-class classification techniques rely more heavily on abnormal samples. If the generated abnormal samples are of poor quality, the method's performance will be severely compromised. As demonstrated in Table~\ref{tab:one_class_classification_summary}, with the exception of MemSeg~\cite{yang2022memseg}, the training of other methods relies on SVDD and Cross-Entropy loss; consequently, the performance of the vast majority of methods is marginally inadequate.

\begin{table}[ht]
\centering
\caption{A summary of one-class classification methods regarding loss function, pre-trained model, and highlights.}
\renewcommand{\arraystretch}{1.2}
\resizebox{\textwidth}{!}{
\begin{tabular}{l p{3cm}<{\raggedright} p{2cm} p{10cm}}
\hline
 \rowcolor{NavyBlue!10} \textbf{Method}      & \textbf{Loss Function }   & \textbf{Pre-trained}       & \textbf{Highlights}       \\ \hline  
Patch SVDD~\cite{yi2020patch} & Cross-Entropy, SVDD           & -        & The paper divides image into patches and sends them to SVDD for training.                           \\
DDSPSVDD~\cite{zhang2021anomaly}    & $L_2$, SVDD                      & VGG                          & DSPSVDD takes reconstruction error into model training.               \\
SE-SVDD~\cite{hu2021semantic}    & SVDD & ResNet                       & The paper proposes a Semantic Correlation module (SCB) to represent abnormal semantics information.   \\
MOCCA~\cite{massoli2021mocca}      & $L_2$, SVDD                      & -            & The paper extends a single boundary to a hard boundary and a soft boundary, it also trains AE as feature extractor.                     \\
        \cite{sauter2021defect}   & Cross-Entropy                       & Xception~\cite{chollet2017xception}                     & The paper uses Xception to train a classification network.            \\
PANDA~\cite{reiss2021panda}      & SVDD, Log-Likelihood                   & DN2~\cite{bergman2020deep} & The paper introduces a method to avoid combating collapse in model adaptation.                            \\
           \cite{sohn2020learning} & Cross-Entropy, Contrastive     & ResNet                       & The paper presents a novel distribution-augmented contrastive learning to enhance the representing ability of network.            \\
           \cite{bai2014saliency} & - & - &  \reviewer{The paper performs template matching on salient regions to detect anomalies.} \\
 \cite{niu2020unsupervised} & $L_1$, $L_2$ & - & \reviewer{This paper uses saliency detection to obtain object contours to assist anomaly detection.} \\
 UISDI~\cite{qiu2020uneven} & $L_1$, $L_2$, Log-Likelihood & - & \reviewer{The paper uses salient object detection to segment the foreground and foreground to obtain abnormal regions.} \\
CutPaste~\cite{li2021cutpaste}   & Cross-Entropy                       & EfficientNet                 & The paper applies ``cut and paste" augmentation into binary anomaly classification.                  \\
         \cite{yoa2021self}  & Cosine Similarity, Contrastive & -         & The paper applies some dynamic local augmentation to generate negative samples.                     \\
CPC-AD~\cite{de2021contrastive}     & InfoNCE                            & -            & The paper applies Contrastive Predictive Coding (CPC) model to AD and get an anomaly score through pixel-wise loss.      \\
MemSeg~\cite{yang2022memseg}     & $L_1$, Focal                      & ResNet                       & The paper artificially creates anomalies in the foreground of products and makes detecting artificial anomalies a segmentation task. \\ \hline
\end{tabular}  
}
\label{tab:one_class_classification_summary}
\end{table}

\begin{figure}[ht]
\centering
    \includegraphics[width=0.75\linewidth]{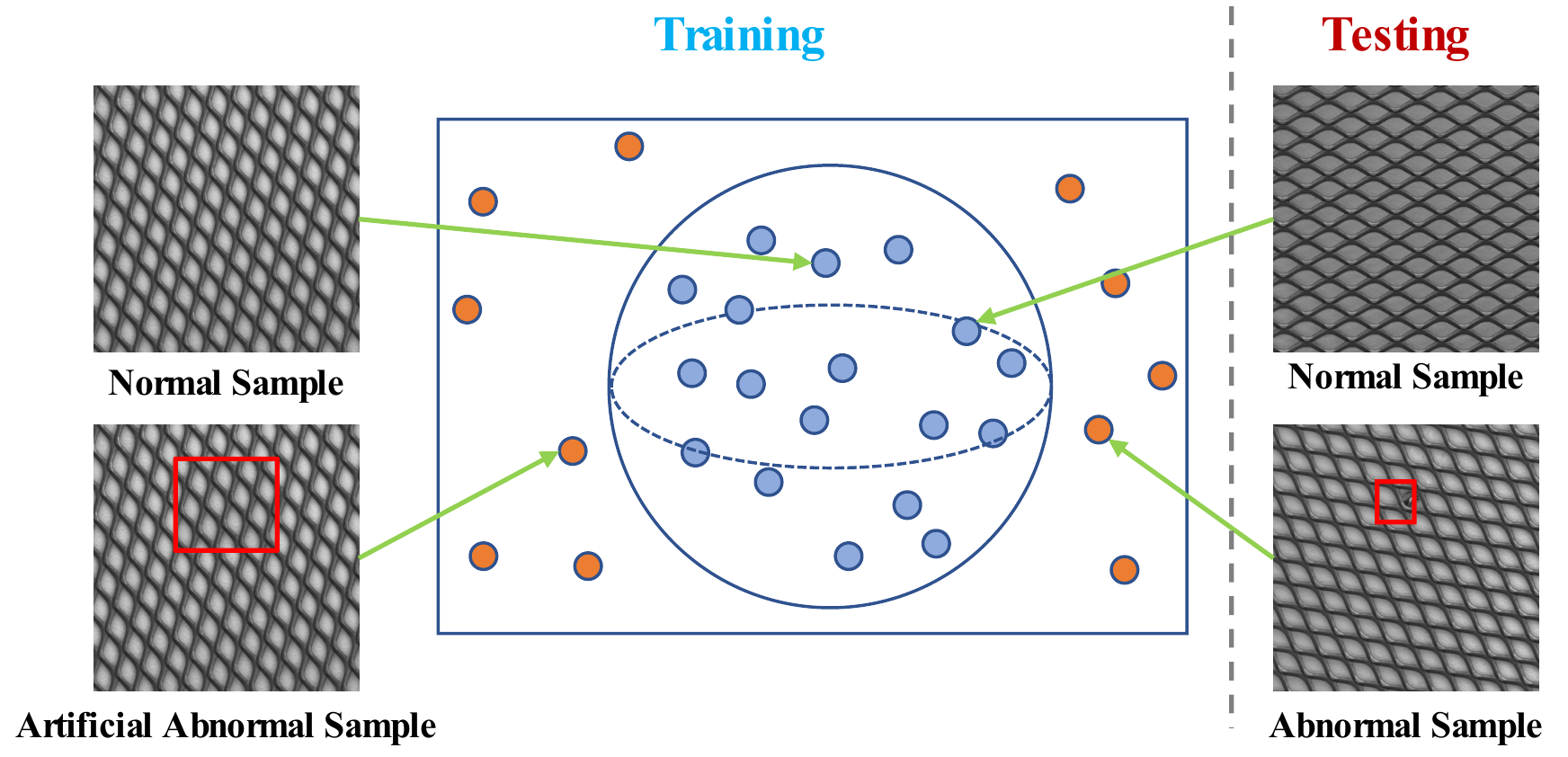}
    \caption{Architecture of one-class classification models.}
\label{fig:OCC_fig}	
\end{figure}

Anomaly detection can also be viewed as a One-Class Classification (OCC) problem, which has inspired some research. As depicted in Fig.~\ref{fig:OCC_fig}, the method finds a hypersphere to distinguish normal sample features from abnormal sample features during training. During inference, the method determines whether the sample is abnormal based on the relative position of the test sample's features and the hypersphere. Since the training set does not contain abnormal samples, some methods create abnormal samples artificially to improve the accuracy of the hypersphere.

SVDD~\cite{tax2004support} is a classic algorithm in the OCC problem, PatchSVDD~\cite{yi2020patch} DSPSVDD~\cite{zhang2021anomaly} and SE-SVDD~\cite{hu2021semantic} improve it for industrial image AD. PatchSVDD~\cite{yi2020patch} divides the image into uniform patches and sends them to the model for training, which significantly enhances the model's ability to detect anomalies. DSPSVDD~\cite{zhang2021anomaly} designs an improved comprehensive optimization objective for the deep SVDD model that simultaneously considers hypersphere volume minimization and network reconstruction error minimization to extract deep data features more effectively. SE-SVDD proposes a Semantic Correlation module (SCB) to improve the representation of abnormal semantics and the accuracy of anomaly localization by extracting multi-level features.

MOCCA~\cite{massoli2021mocca} employs multi-layer features for anomaly detection. MOCCA, unlike SE-SVDD, uses an autoencoder to extract features and locates the boundary position of normal features at each layer. And Sauter~\textit{et al.}~\cite{sauter2021defect} attempt to use the Xception network for classification and obtained results comparable to SVDD. FCDD~\cite{liznerski2020explainable} employs a fully convolutional neural network for OCC. Since the relative positions of the features of each image layer do not change during the convolution process, FCDD yields more interpretable results than alternative methods.

PANDA~\cite{reiss2021panda} examines the migration method of pre-trained features and introduces the early stopping mechanism to the OCC problem. In addition, Reiss~\textit{et al.}~\cite{reiss2021mean} investigate the issue of catastrophic forgetting in PANDA. They propose a new loss function capable of overcoming the failure modes of both center-loss and contrastive-loss methods and replacing Euclidean distance with a confidence-invariant angular center loss for prediction.
	
DisAug CLR~\cite{sohn2020learning} proposes a two-stage anomaly detection framework, in which the first stage hinders the uniformity of contrastive representations by means of a novel distribution-enhanced contrastive learning. After comparative learning, abnormal and normal sample representations are easier to distinguish. While the second stage builds a one-class classifier using the representations learned in the first stage. Yoa \textit{et al.}~\cite{yoa2021self} presents a novel dynamic local augmentation to generate negative image pairs from a normal training dataset, which is effective for anomaly detection. Contrastive Predictive Coding (CPC) ~\cite{oord2018representation} model is utilized by De \textit{et al.}~\cite{de2021contrastive} for anomaly detection and segmentation, which uses patch-wise contrastive loss as anomaly score to localize anomalies.


\reviewer{In addition, inspired by saliency object detection~\cite{zhang2019synthesizing,zhuge2022salient,fang2022densely}, many methods apply saliency detection to anomaly detection.. Bai et     al.~\cite{bai2014saliency} proposed to use Fourier transform to detect salient regions of images, and compare the salient regions with templates to detect anomalies. Niu et al.~\cite{niu2020unsupervised} used the method of salient object detection to obtain object contours, thereby assisting the detection of outliers. Qiu et al.~\cite{qiu2020uneven} proposed a Multi-Scale Saliency Detection (MSSD) method to separate the foreground and foreground to obtain coarse anomaly regions, and refine the detected results on this basis. What's more, GradCAM~\cite{selvaraju2017grad}, as a common method to obtain saliency maps, is also used in various anomaly detection algorithms. Both CutPaste~\cite{li2021cutpaste} and CAVGA~\cite{venkataramanan2020attention} treat anomaly detection as a classification problem, while GradCAM is used for pixel-level anomaly localization.}

CutPaste~\cite{li2021cutpaste} is a representative example of an OCC method for data augmentation. It generates abnormal images by cutting and pasting portions of normal images, allowing the network to distinguish abnormal images. Additionally, segmentation-based methods are useful. This method puts more emphasis on pixel-level anomaly localization. When the flow is known, Iquebal~\textit{et al.}~\cite{iquebal2020consistent} demonstrate that the maximum posterior estimation of image labels can be formulated as a continuous max-flow problem. Then, anomaly segmentation is accomplished by obtaining flows iteratively using a novel Markov random field on the image domain. The technique shows its adaptability using a dataset for metal additive manufacturing anomaly detection~\cite{attar2014manufacture}. MemSeg~\cite{yang2022memseg} stores the features of normal images in a memory bank in order to improve the segmentation network's ability to distinguish abnormal regions. In order to prevent the influence of background factors, MemSeg only introduces anomalies in external data sets in the foreground of items, which is another reason for its excellent performance.

\subsubsection{Distribution Map}

Distribution-map based methods necessitate a suitable mapping objective for training, and the choice of mapping method impacts model performance. As shown in Table~\ref{tab:distribution-map-based_summary}, Normalizing Flows (NF)-based methods predominate. As a generative model, NF has a strong mapping ability, and it has also demonstrated good performance in AD tasks.

\begin{table}[ht]
\centering
\caption{A summary of distribution-map based methods regarding loss function, pre-trained model, and highlights.}
\renewcommand{\arraystretch}{1.2}
\resizebox{\textwidth}{!}{
\begin{tabular}{l p{3cm}<{\raggedright} p{2cm} p{10cm}}
\hline
 \rowcolor{NavyBlue!10} \textbf{Method}      & \textbf{Loss Function }   & \textbf{Pre-trained}       & \textbf{Highlights}       \\ \hline  
         \cite{tailanian2021multi} & PCA              & ResNet               & The paper uses PCA and ResNet to extract features and count their distribution.                              \\
         \cite{rippel2021modeling} & Cross-Entropy      & ResNet, EfficientNet & The paper establishes a model of normality by fitting a multivariate Gaussian to feature representations of a pre-trained network.         \\
         \cite{rippel2021transfer} & Mahalanobis Distance       & EfficientNet         & The paper generates a multi-variate Gaussian distribution for the normal class and mitigates the catastrophic forgetting in past research. \\
PEDENet~\cite{zhang2022pedenet}   & Log-Likelihood, Cross-Entropy, Regularization  & -     & The model can predict the location of the patch and compare it with the actual location to judge the abnormality.                          \\
PFM~\cite{wan2022unsupervised}       & $L_2$                         & ResNet               & The paper proposes the bidirectional and multi-hierarchical bidirectional pre-trained feature mapping based on the vanilla feature mapping.      \\
PEFM~\cite{wan2022position}      & $L_2$                         & ResNet               & The paper introduces position encoding into PFM.      \\
FYD~\cite{zheng2022focus}       & $L_2$                         & ResNet               & The paper aligns samples at image and feature levels to detect anomalies.            \\
DifferNet~\cite{rudolph2021same} & Log-Likelihood             & ResNet               & The paper is the first one to introduce normalizing flow into anomaly detection.                           \\
CS-Flow~\cite{rudolph2022fully}   & Log-Likelihood             & ResNet               & The paper uses information of multi-scale feature maps and improves DifferNet.    \\
CFlow-AD~\cite{gudovskiy2022cflow}  & Log-Likelihood             & ResNet               & The paper introduces positional encoding into the conditional normalizing flow framework.                     \\
CAINNFlow~\cite{yan2022cainnflow} & Log-Likelihood             & ViT~\cite{dosovitskiy2020image}                  & The paper uses VIT to replace ResNet and achieve better result.                \\
FastFlow~\cite{yu2021fastflow}  & Log-Likelihood             & ResNet               & The paper introduces an alternate stacking of large and small convolution kernels in the NF module to model global and local distribution.  \\
AltUB~\cite{kim2022altub}   & Log-Likelihood             & ResNet               & The paper designs a module for normalizing flow based methods and improve their performance. \\ \hline
\end{tabular}  
}
\label{tab:distribution-map-based_summary}
\end{table}

\begin{figure}[t]
    \centering
    \includegraphics[width=0.75\linewidth]{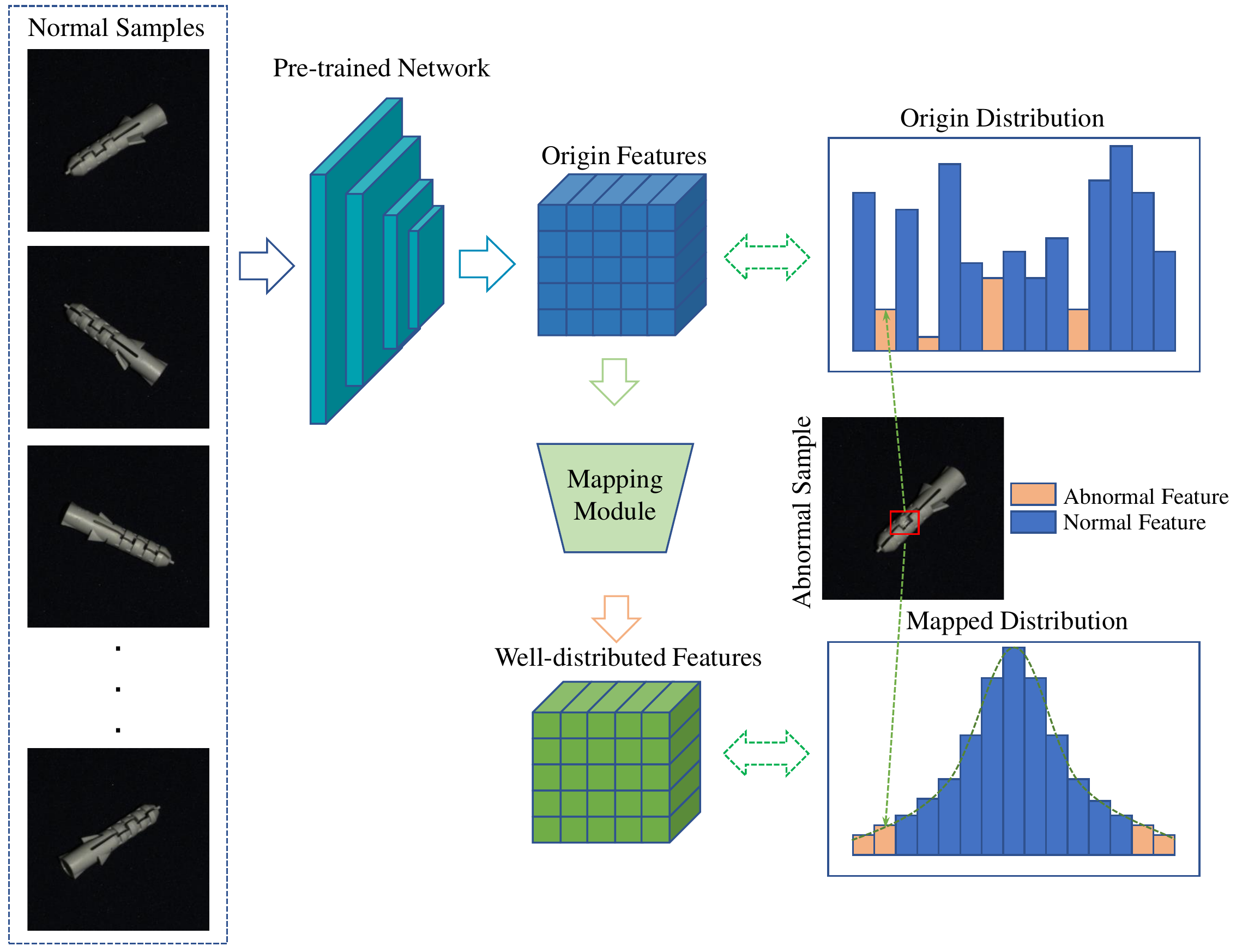}
    \caption{Architecture of distribution-map based methods.}
    \label{fig:mapping_fig}
\end{figure}

Distribution-map based methods are very similar to OCC-based methods, with the exception that OCC-based methods concentrate on finding feature boundaries, whereas mapping-based methods attempt to map features into desired distributions. A common framework for those methods is shown in Fig.~\ref{fig:mapping_fig}. This expected distribution is typically a MultiVariate Gaussian (MVG) distribution. This type of method first employs a strong pre-trained network to extract the features of normal images, and then maps the extracted features to the Gaussian distribution using a mapping module. This distribution will be deviated from by the features of abnormal images that appear during the evaluation. The abnormal probability can be calculated based on the level of deviation.

Tailanian \textit{et al.}~\cite{tailanian2021multi} propose a contrario framework that applies statistical analysis to feature maps produced by patch PCA and ResNet, which performs well on leather samples, to detect anomalies in images. By fitting a multivariate Gaussian to the feature representations of a pre-trained network, Rippel \textit{et al.}~\cite{rippel2021modeling} establish a model of normality. Nonetheless, the issue of catastrophic forgetting remains unresolved. Based on the relationship between generative and discriminative modeling, Rippel \textit{et al.}~\cite{rippel2021transfer} generate a multi-variable Gaussian distribution for the normal class and prove the efficacy of this concept on Deep SVDD and FCDD, which mitigates the catastrophic forgetting observed in previous research. PEDENet~\cite{zhang2022pedenet} framework consists of a Patch Embedding (PE) network, a Density Estimation (DE) network, and a Location Prediction (LP) network. At first, the PE module is used to reduce the size of the features that the pre-trained network has extracted. Then, using the DE module, which was inspired by the Gaussian mixture model, and the LP module, the model can predict the relative position of the patch embedding and, based on the difference between the predicted result and the actual result during inference, decide if the image is abnormal. Pre-trained Feature Mapping (PFM)~\cite{wan2022unsupervised} proposes bidirectional and multi-hierarchical bidirectional pre-trained feature mapping to enhance the performance of vanilla feature mapping. In addition, Wan~\textit{et al.} \cite{wan2022position} add position encoding to the PFM framework and propose a novel Position Encoding enhanced Feature Mapping (PEFM)~\cite{wan2022position} to further enhance PFM. FYD~\cite{zheng2022focus} introduces registration to industrial image AD for the first time. FYD suggests a coarse-to-fine alignment method that starts with aligning the foreground of objects at the image level. Next, in the refinement alignment stage, non-contrastive learning is used to increase the similarity of features between all corresponding positions in a batch.

Normalizing Flows (NF)~\cite{JimenezRezende2015VariationalIW} is a technique for constructing complex distributions by transforming a probability density via a series of invertible mappings. NF methods extract features from normal images from a pre-trained model, such as ResNet~\cite{He2015DeepRL} or Swin Transformer~\cite{Liu2021SwinTH}, and transform the feature distribution as a Gaussian distribution during the training phase. In the test phase, after passing through NF, the features of abnormal images will deviate from the Gaussian distribution of the training phase, which is the most important principle for classifying anomalies. DifferNet~\cite{rudolph2021same} is the first research to use NF to address the industrial image AD issue. By incorporating cross-convolution blocks within the normalizing flow to assign probabilities, CS-Flow~\cite{rudolph2022fully} makes use of the context within and between multi-scale feature maps to improve DifferNet. CFlow-AD~\cite{gudovskiy2022cflow} adds positional encoding to the framework for conditional normalizing flow to achieve superior results. In addition, CFlow-AD~\cite{gudovskiy2022cflow} analyzes in depth why the multivariate Gaussian assumption is a reasonable prior in earlier models and why the more general NF framework aims to converge to similar results with less computation. FastFlow~\cite{yu2021fastflow} introduces an alternate stacking of large and small convolution kernels in the NF module to model global and local distribution efficiently. CAINNFlow~\cite{yan2022cainnflow} enhances the performance of the model by introducing the attention mechanism CBAM~\cite{woo2018cbam} to the NF module. In techniques such as FastFlow and CFlow-AD, the feature distribution center is not 0 and their performance is unstable. Kim \textit{et al.}~\cite{kim2021semi} propose a simple solution AltUB~\cite{kim2022altub} that uses alternating training to update the base distribution of normalizing flow for anomaly detection in order to solve the problem. AltUB verifies the effect of CFlow-AD and FastFlow using AltUB.

\subsubsection{Memory Bank}

As illustrated in Table~\ref{tab:memory_bank_summary}, memory-based methods regularly do not require the loss function for training, and models are constructed quickly. Their performance is ensured by a robust pre-training network and additional memory space, and this type of method is currently the most effective in IAD tasks.

\begin{table}[ht]
\centering
\caption{A summary of memory bank based methods regarding loss function, pre-trained model, and highlights.}
\renewcommand{\arraystretch}{1.2}
\resizebox{\textwidth}{!}{
\begin{tabular}{l p{3cm}<{\raggedright} p{2cm} p{10cm}}
\hline
 \rowcolor{NavyBlue!10} \textbf{Method}      & \textbf{Loss Function }   & \textbf{Pre-trained}       & \textbf{Highlights}       \\ \hline  
SPADE~\cite{cohen2020sub}     & -      & ResNet            & The paper uses multi-resolution feature to detect anomalies based on KNN.                      \\
\cite{kim2021semi}          & -        & ResNet            & The paper reduces the computational cost for the inverse of multi-dimensional covariance tensor so that bigger resolution image can be applied. \\
SOMAD~\cite{li2021anomaly}     & -        & ResNet            & The paper maintains normal characteristics by using topological memory based on multi-scale features.                 \\
GCPF~\cite{wan2021industrial}      & -        & ResNet            & The paper processes normal features into multiple independent multivariate Gaussian clustering.                        \\
MSPB~\cite{tsai2022multi}      & Kmeans, Cosine Similarity, SVDD   & VGG               & The paper enhances network representation capabilities by learning patch position relationships.                           \\
SPD~\cite{zou2022spot}       & Focal, InfoNCE, SPD, Cosine Similarity & -  & Design a contrastive learning method to retrain ResNet to enhance the ability of defect representation.           \\
PatchCore~\cite{roth2022towards} & -        & ResNet            & The paper introduces a core-set sampling method to build a memory bank.                     \\
CFA~\cite{Lee2022CFACF}       & SVDD               & ResNet            & The paper improves PatchCore so that image features are distributed on a hypersphere.          \\
FAPM~\cite{kim2022fapm}      & -        & ResNet            & The paper puts different position features of the image into different memory banks to speed up retrieval.                   \\
N-pad~\cite{bae2022image}     & Mahalanobis Distance, Log-Likelihood                  & ResNet            & The paper allows for possible edge misalignment by estimating a nominal distribution for each pixel using the pixel's neighborhood features.
\\ \hline
\end{tabular}  
}
\label{tab:memory_bank_summary}
\end{table}

\begin{figure}[t]
\centering
    \includegraphics[width=0.85\linewidth]{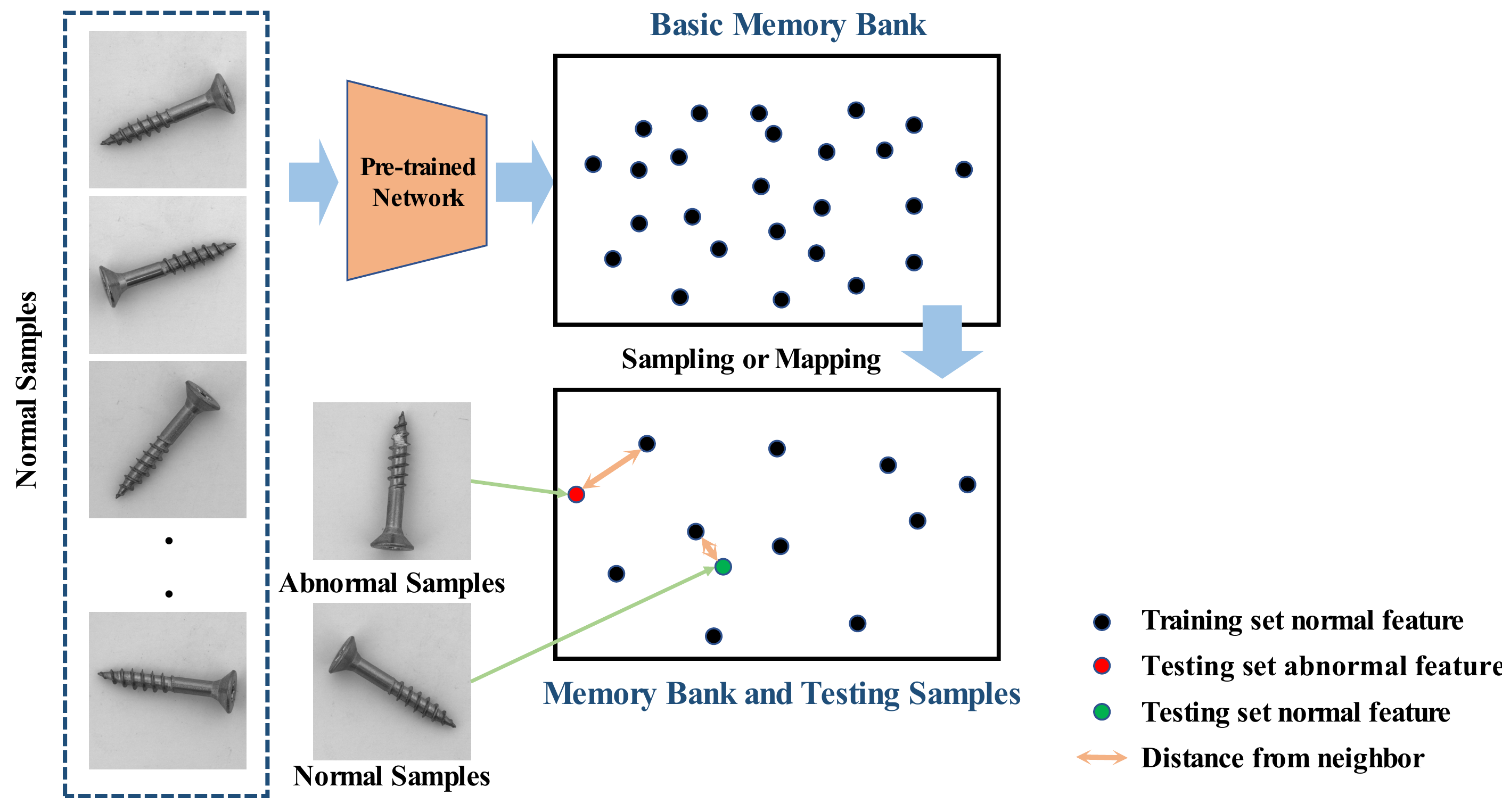}
    \caption{Architecture of memory bank based methods.}
    \label{fig:memory_bank}
\end{figure}

The primary distinction between memory bank-based methods and OCC-based methods, is that memory-based methods, such as SVDD, require additional memory space to store image features. As shown in Fig.~\ref{fig:memory_bank}, these methods require minimal network training and only require sampling or mapping the collected normal image features for inference. During inference, features of the test image are compared to features in the memory bank. The abnormal probability of the test image is equal to the spatial distance from the normal features in the memory bank.

K Nearest Neighbors (KNN)~\cite{Eskin2002AGF} is a widely used algorithm for unsupervised anomaly detection, but it operates only at the sample level. Semantic Pyramid Anomaly Detection (SPADE)~\cite{cohen2020sub} is inspired by KNN and utilizes correspondences based on a multi-resolution feature pyramid to obtain pixel-level anomaly segmentation results. PaDim~\cite{defard2021padim} employs multivariate Gaussian distributions to construct a probabilistic representation of the normal class. Consequently, the memory bank size is determined solely by the image resolution and not by the size of the training set. PaDiM requires the batch-inverse of the multidimensional covariance tensor, which makes it challenging to scale up to larger CNNs due to the increased feature size. To reduce the computational cost of the inverse by a factor of three, Kim \textit{et al.}~\cite{kim2021semi} generalize random feature selection into semi-orthogonal embedding.

Meanwhile, Self-organizing Map for Anomaly Detection (SOMAD)~\cite{li2021anomaly} and GCPF~\cite{wan2021industrial} enhance the storage of normal features. SOMAD preserves normal characteristics by employing topological memory based on multi-scale features. While GCPF transforms standard characteristics into multiple independent multivariate Gaussian clustering.

PatchCore~\cite{roth2022towards} is a significant advancement in industrial image AD that significantly raises the performance for MVTec AD. PatchCore contains two special points. First, the memory bank of PatchCore is coreset-subsampled to ensure a low inference cost while maximizing performance. PatchCore then determines whether the test sample is abnormal based on the distance between the test sample's nearest neighbor feature in its memory bank and other features. This process of reweighting makes PatchCore more robust. Since PatchCore was proposed, numerous improved methods have been developed on its foundation. Coupled-hypersphere-based Feature Adaptation (CFA) is proposed by Lee~\textit{et al.}~\cite{Lee2022CFACF} to obtain target-oriented features. The center and surface of the hypersphere in the memory bank are obtained through transfer learning, and the positional relationship between the test feature and the coupled-hypersphere can be used to determine whether it is abnormal or not. FAPM~\cite{kim2022fapm} is comprised of numerous patch-wise and layer-wise memory banks located in various places. FAPM calculates the features in different memory banks independently during inference, which significantly accelerates inference speed. N-pad~\cite{jang2022n} allows for the possibility of marginal misalignment by estimating a per-pixel nominal distribution using neighboring and target pixel features. In addition, anomaly scores are deduced using both Mahalanobis and Euclidean distances between target pixels and the estimated distribution. Similarly, Bae~\textit{et al.}~\cite{bae2022image} model the cumulative histogram using location information as conditional probabilities, and neighborhood information was used to establish the normal feature distribution. Furthermore, this work introduces the first refinement approach in the anomaly detection and localization problem, using synthetic anomalous images to improve the anomaly map based on the input image, as well as using neighborhood and location information to estimate the distribution.

By learning the embedding position information and comparing the extracted features with the normal embedding during inference, Tsai~\textit{et al.}~\cite{tsai2022multi} propose a method to improve the network's ability to represent data. It is also based on the concept of self-supervised learning. Zou~\textit{et al.}~\cite{zou2022spot} use contrastive learning to train the backbone network and propose a new data augmentation method called SPD to push the network to differentiate between two images with slight differences. In addition, they demonstrate the representation capability of the backbone network using PatchCore~\cite{roth2022towards}.

\begin{table}[!ht]
\caption{A summary of reconstruction based methods.}
\renewcommand{\arraystretch}{1.2}
\resizebox{\textwidth}{!}{
\begin{tabular}{p{2cm} p{3cm}<{\raggedright} p{2cm} p{10cm}}
\hline
 \rowcolor{NavyBlue!10} \textbf{Method}      & \textbf{Loss Function}   & \textbf{Pre-trained}       & \textbf{Highlights}       \\ \hline  
   \rowcolor{gray!10} \multicolumn{4}{c}{ \textcolor{NavyBlue}{(1) Autoencoder Model}} \\ \hline
   
     \cite{bergmann2018improving}      & $L_2$, SSIM                         & -  & The paper firstly takes SSIM as a loss to reconstruct image and detect anomalies.              \\
       \cite{chung2020unsupervised}     & $L_2$, SSIM                         & -  & The paper proposes two AEs and reduces style change during image reconstruction.               \\
     UTAD~\cite{liu2021unsupervised}    & $L_1$, Adversarial                  & VGG               & The paper uses two-stage reconstruction to generate high-fidelity images to avoid reconstruction errors.                    \\
     DFR~\cite{yang2020dfr}     & $L_2$      & VGG               & The paper proposes to reconstruct and compare at the feature level to detect anomalies.              \\
     ALT~\cite{yan2021unsupervised}     & $L_1$, Perceptual, Adversarial & VGG               & The paper proposes an adaptive attention-level transition strategy and uses perceptual loss to improve reconstruction quality.  \\
     P-Net~\cite{zhou2020encoding}   & $L_1$, Adversarial                  & -  & The paper designs a new architecture for anomaly detection.                       \\
     \cite{collin2021improved}        & $L_1$, $L_2$                          & -  & The paper adds skip-connection in reconstruction network and adds noise during training to improve reconstruction sharpness. \\
        \cite{tao2022unsupervised}     & $L_2$      & VGG               & The paper proposes a dense feature fusion module to assist reconstruction.                     \\
        \cite{hou2021divide}     & $L_2$, Adversarial                  & -  & The paper uses memory to help reconstructing images.             \\
     EdgRec~\cite{liu2022reconstruction}  & $L_2$, SSIM                         & -  & The paper reconstructs from the gray value edge and preserves the high-frequency information with skip-connection.           \\
    PAE~\cite{kim2022spatial}     &  $L_2$, Cross-Entropy               & -  & The paper gradually increases the resolution of the input image during training.                     \\
    SMAI~\cite{li2020superpixel}    &  $L_2$, SSIM      & -  & The paper masks and inpaintings image by superpixel.                  \\
    RIAD~\cite{zavrtanik2021reconstruction}    & $L_2$, MSGMS, SSIM             & -  & The paper proposes to inpaint and reconstruct images by patch.    \\
    I3AD~\cite{nakanishi2020iterative}    & $L_1$, Adversarial                  & -  & The paper gradually masks the high anomaly probability areas and reconstructs them.                 \\
        \cite{bauer2022self}    & $L_2$      & -  & The paper proposes to reconstruct the anomalous area differently from the original image.      \\
       \cite{huang2022self}     & $L_2$, SSIM, GMS             & -  & Similar to I3AD, but the paper adds skip connections to reconstruction network.               \\
    DREAM~\cite{zavrtanik2021draem}   & $L_2$, SSIM, Focal             & -  & The paper designs a method to generate abnormal images and uses U-Net~\cite{ronneberger2015u} to distinguish anomalies after reconstruction. \\
    SGSF~\cite{xing2022self} & $L_2$, SSIM, Focal & - & The method utilizes the idea of saliency detection to generate more realistic anomalies than DRAEM.          \\
    DSR~\cite{zavrtanik2022dsr}     & $L_2$, Focal                        & -  & The paper generates abnormal samples in feature level and perform better than DRAEM.                 \\
    NSA~\cite{schluter2022natural}     & $L_2$, Cross-Entropy               & -  & The paper generates abnormal samples by pasting parts of other normal samples, which is the SOTA method without extra data.        \\
    SSPCAB~\cite{ristea2022self}  & $L_2$      & -  & The paper designs a “plug and play” self-supervised block to improve the reconstruction ability of many methods.             \\
    SSMCTB~\cite{madan2022self}  & $L_2$ & - & This paper replaces the SE-layer in SSPCAB with transformer architecture.     \\
      \cite{dehaene2019iterative}     & Cross-Entropy                         & -  & The paper guides reconstruction using gradient descent with VAE.                               \\
       \cite{liu2020towards}     & Attention Disentanglement             & -  & The paper proposes to use disentanglement VAE to detect anomalies.                            \\
    DGM~\cite{matsubara2020deep}     & $L_2$, Log-Likelihood               & -  & The paper proposes to use non-regularized objective functions for training VAE under heterogeneous datasets.                 \\
    FAVAE~\cite{dehaene2020anomaly}   & Log-Likelihood    & VGG               & The paper uses VAE to model the distribution of features extracted by its pre-trained model.       \\
       \cite{wang2020image}     & $L_2$, Cross-Entropy                & -  & The paper uses VQ-VAE to construct a discrete latent space and reconstructs images based on the latent space.                \\ \hline  
    
    \rowcolor{gray!10} \multicolumn{4}{c}{\ \textcolor{NavyBlue}{(2) GAN Model}} \\ \hline
    
     SCADN~\cite{yan2021learning}   & $L_2$, Adversarial                 & -  & The paper masks part of image and reconstruct image with GAN during training.                      \\
    AnoSeg~\cite{song2021anoseg}  & $L_1$, $L_2$, Adversarial         & -  & The paper generates abnormal samples through a GAN and detects anomalies with the discriminator.  \\
    OCR-GAN~\cite{liang2022omni} & $L_1$, $L_2$, Adversarial         & -  & The paper uses the Frequency Decoupling module to decouple and reconstruct images.          \\ \hline  
    
    \rowcolor{gray!10} \multicolumn{4}{c}{ \textcolor{NavyBlue}{(3) Transformer Model}} \\  \hline
    
    VT-ADL~\cite{mishra2021vt}  & $L_2$, SSIM, Log-Likelihood    & -  & The paper proposes a transformer-based framework to reconstruct images and detects anomalies.  \\
    ADTR~\cite{you2022adtr}    & $L_2$, Cross-Entropy                & EfficientNet      & The paper makes it simple to identify anomalies when reconstruction fails by reconstructing features from pre-trained network.    \\
    AnoViT~\cite{lee2022anovit}  & $L_2$      & ViT               & The paper uses a pre-trained ViT to extract features and reconstruct images.                      \\
    HaloAE~\cite{mathian2022haloae}  & $L_2$, Cross-Entropy, SSIM     & VGG               & The paper introduces an auto-encoder architecture based on a transformer with HaloNet.         \\
    InTra~\cite{pirnay2022inpainting}   & $L_2$, GMS, SSIM               & -  & The paper leverages more global information to repair images with transformer.           \\
    MSTUnet~\cite{jiang2022masked} & \makecell[l]{$L_2$, SSIM, Focal}             & -  & The paper uses swin transformer for inpainting masked images and detects anomalies.            \\
    MeTAL~\cite{de2022masked}   & $L_1$, SSIM                         & -  & The paper uses information from neighbor patches to inpainting images, better accounting for local structural information.       \\
    UniAD~\cite{you2022unified}   & $L_2$      & EfficientNet      & The paper trains all categories of products in one model.                           \\ \hline
    
    \rowcolor{gray!10} \multicolumn{4}{c}{ \textcolor{NavyBlue}{(4) Diffusion Model}} \\ \hline
    
    AnoDDPM~\cite{wyatt2022anoddpm} & $L_2$, Log-Likelihood               & -  & The paper is the first to apply diffusion model for industrial image anomaly detection.        \\
     \cite{teng2022unsupervised}    & $L_2$, Log-Likelihood               & -  & The paper significantly speeds up the inference process of anomaly detection using diffusion model.                 \\ \hline        
\end{tabular}}
\label{tab:recontruction_based_summary}
\end{table}


Reconstruction-based methods primarily self-train encoders and decoders to reconstruct images for anomaly detection, which makes them less reliant on the pre-trained model and increases their ability to detect anomalies. However, its image classification capability is poor due to its inability to extract high-level semantic features. As shown in Table~\ref{tab:recontruction_based_summary}, the loss functions of various methods are comparable; however, their performance varies due to different reconstruction model paradigms and abnormal sample construction methods.

\begin{figure}[h]
\centering
    \includegraphics[width=0.75\linewidth]{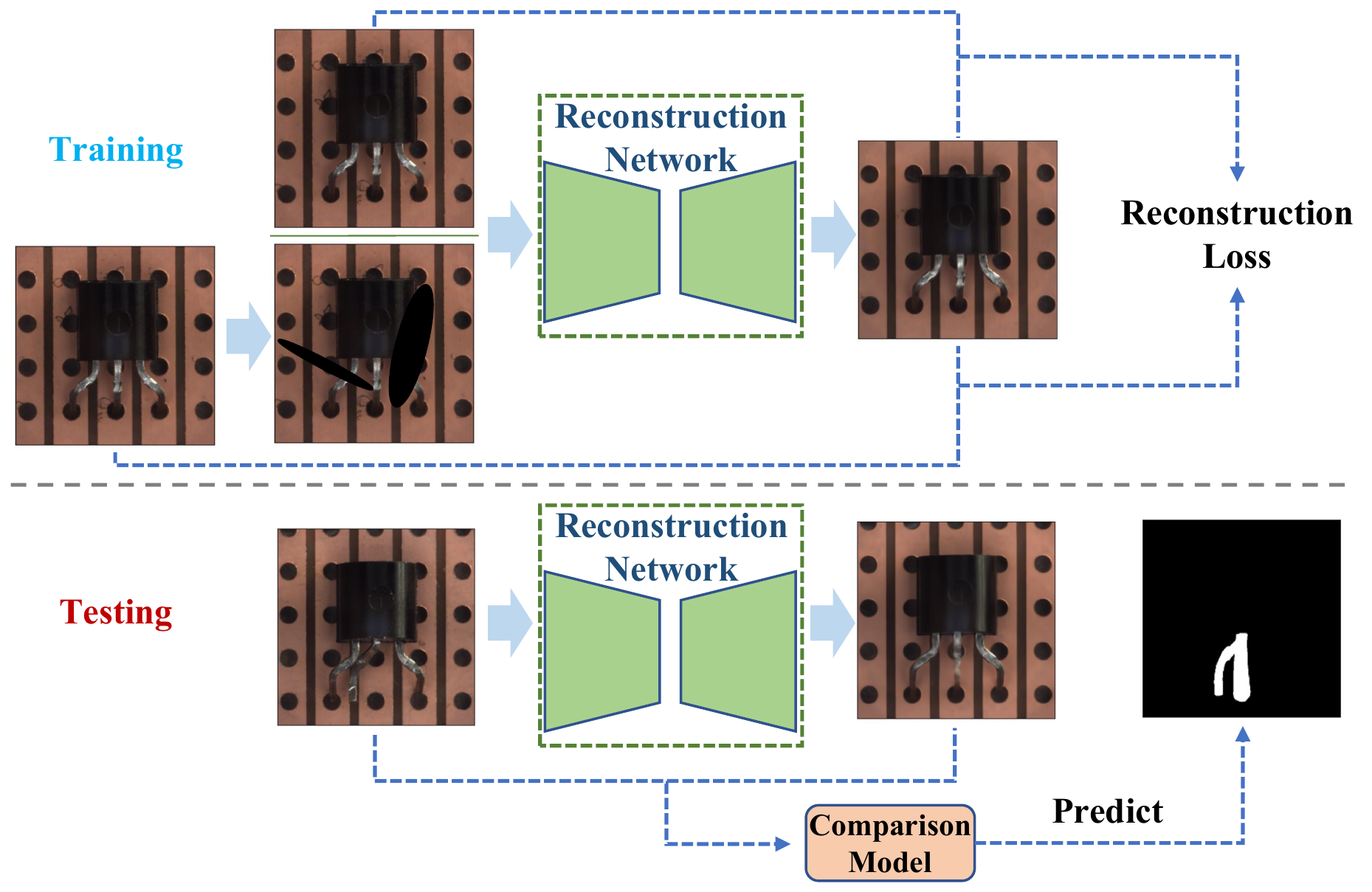}
    \caption{Architecture of reconstruction based models.}
    \label{fig:reconstruction}
\end{figure}

The structure of the reconstruction-based technique is depicted in Fig.~\ref{fig:reconstruction}. During the training process, normal or abnormal images are sent to the reconstruction network, and the reconstruction loss function is used to guide the training of the reconstruction network. Finally, the reconstruction network can restore the reconstruction image in a manner similar to the original normal image. In the inference stage, the comparison model compares the original image to the reconstructed image to generate a prediction. In contrast to the variety of methods for feature embedding, the majority of reconstruction-based methods only differ in the construction of the reconstruction network. Reconstruction-based methods outperform feature-embedding methods at the pixel level due to their ability to identify anomalies through pixel-level comparison. In addition, the majority of reconstruction-based methods are trained from scratch without employing robust pre-trained models, which results in inferior performance compared to image-level feature embedding.

\subsubsection{Autoencoder}
Autoencoder (AE) is the most prevalent reconstruction network for AD. Numerous other reconstruction networks also consist of encoder and decoder components. Bergmann~\textit{et al.}~\cite{bergmann2018improving} investigate the influence of Structure Similarity Index Measure (SSIM) and $L_2$ loss on AE reconstruction and anomaly segmentation, providing numerous suggestions for future research.

How to resolve the difference between the reconstructed image and the original image is the most foundational principle. There are regularly differences in style between the reconstructed image and the original image, resulting in over-detection. Chung~\textit{et al.}~\cite{chung2020unsupervised} present an Outlier-Exposed Style Distillation Network (OE-SDN) to preserve the style translation and suppress the content translation of the AE in order to avoid over-detection. As the anomaly prediction, Chung~\textit{et al.} replaced the difference between the original image and the reconstruction image of AE with the difference between the reconstruction image of OE-SDN and the reconstruction image of AE. Unsupervised Two-stage Anomaly Detection (UTAD)~\cite{liu2021unsupervised} brings an IE-Net and Expert-Net to extract and utilize impressions for anomaly-free and high-fidelity reconstructions, thereby offering the framework interpretable.

Reconstruction-based methods are nearly effective as feature embedding methods when utilizing features at different scales. Similar to teacher-student architecture, Deep Feature Reconstruction (DFR)~\cite{yang2020dfr} method detects anomalous through reconstruction at the level of features. DFR obtains multiple spatial context-aware representations from a network that has been pre-trained. Then, DFR reconstructs features using a deep yet efficient convolutional AE and detects anomalous regions by comparing the original features to the reconstruction features. Yan~\textit{et al.}~\cite{yan2021unsupervised} propose a novel Multi-Level Image Reconstruction (MLIR) framework that forms the reconstruction process as an image denoising task at different resolutions. Thus, MLIR accounts for the detection of both global structure anomalies and detail anomalies.

Modifying the structure of AE can also improve its capacity for reconstruction. Zhou~\textit{et al.}~\cite{zhou2020encoding} introduce P-Net to compare the difference in structure between the original and reconstruction images. Collin~\textit{et al.}~\cite{collin2021improved} include skip-connections between encoder and decoder to improve the reconstruction's sharpness. In addition, they propose corrupting them with a synthetic noise model to prevent the network from convergently mapping identities, and they introduce the innovative Stain noise model for this purpose. Tao~\textit{et al.}~\cite{tao2022unsupervised} also operate at the feature level; they employ a dense feature fusion module to obtain a dense feature representation of double input in order to help reconstruction in the dual-Siamese framework. Hou~\textit{et al.}~\cite{hou2021divide} also use skip-connections to enhance the quality of reconstruction. In addition to achieving expected results, they add a memory module to skip-connections. Liu~\textit{et al.}~\cite{liu2022reconstruction} reconstruct the original RGB image from its gray value edges, with the skip-connections in the model preserving the image's high-frequency information to better guide the reconstruction. Progressive Autoencoder (PAE)~\cite{kim2022spatial} improves autoencoder reconstruction performance through progressive learning and modified CutPaste augmentation. During training, PAE achieves progressive learning by gradually increasing the input image's resolution.

Masking and repainting is an effective method for self-supervised learning. The Superpixel Masking And Inpainting (SMAI) technique was developed by Li~\textit{et al.}~\cite{li2020superpixel}. SMAI divides the image into multiple blocks of superpixels and trains the inpainting module to reconstruct a superpixel within a mask. SMAI performs masking and inpainting superpixel-by-superpixel on the test image during inference, and then compares the reconstruction image to the test image to distinguish abnormal regions. Iterative Image Inpainting Anomaly Detection (I3AD) is a method proposed by Nakanishi~\textit{et al.}~\cite{nakanishi2020iterative} that reconstructs partial regions based on the anomaly map. I3AD improves reconstruction quality by only reconstructing inpainting masks over images, and only masking regions with a high probability of abnormality. SSM~\cite{huang2022self} is conceptually similar to I3AD. SSM adds skip-connections to the reconstruction network and predicts the mask region as the training target. RIAD~\cite{zavrtanik2021reconstruction} randomly masks a portion of the training set image at the patch level and reconstructs it using a U-Net encoder-decoder network~\cite{ronneberger2015u}. During inference, RIAD combines multiple random masks and reconstruction patches to generate a reconstructed image, which is then compared to the original image. Multi-Scale Gradient Magnitude Similarity (MSGMS) outperforms SSIM as an anomaly score, according to RIAD.

DRAEM~\cite{zavrtanik2021draem} is representative of reconstruction-based techniques. DRAEM synthesizes abnormal images and reconstructs them as normal by introducing external datasets, which greatly improves the reconstruction network's generalization capacity. In addition, DRAEM feeds the original image and the reconstructed image into the segmentation network to predict abnormal regions, significantly enhancing the model's ability to segment anomalous regions. Nevertheless, DRAEM is susceptible to failure when synthesizing near-in-distribution anomalies. \reviewer{Inspired by saliency detection, Xing et al.~\cite{xing2022self} proposed the Saliency Augmentation Module (SAM) to generate more realistic abnormal images than DRAEM, so as to achieve better results.} DSR~\cite{zavrtanik2022dsr} proposes an architecture based on quantized feature space representation and dual decoders to circumvent the requirement for image-level anomaly generation. By sampling the learned quantized feature space at the feature level, the near-in-distribution anomalies are generated in a controlled way. NSA~\cite{schluter2022natural} does not use external data for data data augmentation and adopts more data augmentation methods, allowing it to outperform all previous methods that learned without utilizing additional datasets. In contrast to other methods that attempt to reconstruct abnormal images into normal images, Bauer~\cite{bauer2022self} proposes reconstructing the abnormal areas of the image so that they deviate from the original image's appearance. This approach produces comparable results to other methods.

In contrast to classical reconstruction-based methods, Ristea~\textit{et al.}~\cite{ristea2022self} propose integrating reconstruction-based functionality into a Self-Supervised Predictive Architectural Building Block (SSPCAB). SSPCAB can be incorporated into models such as DRAEM and CutPaste to enhance those models. Self-Supervised Masked Convolutional Transformer Block (SSMCTB)~\cite{madan2022self} transforms the SE-layer~\cite{hu2018squeeze} in SSPCAB into a channel-wise transformer block and achieves superior results.

VAE is a variant of AE, with the difference that the intermediate variables of VAE are data from a normal distribution. Naturally, VAE has superior interpretability. Dehaene~\textit{et al.}~\cite{dehaene2019iterative} iteratively guide reconstruction using gradient descent with energy defined by the reconstruction loss, thereby overcoming the tendency of VAE to produce blurry reconstructions and preserving the normal high-frequency structure. The variational autoencoder is trained with an attention disentanglement loss by Liu~\textit{et al}~\cite{liu2020towards}. Anomaly inputs in this VAE will result in Gaussian-deviating latent variables during gradient backpropagation and attention generation. This deviation can be used to locate anomalies. According to Matsubara~\textit{et al.}~\cite{matsubara2020deep}, datasets are commonly heterogeneous rather than regularized, and non-regularized objective functions are more suitable for training VAE models on heterogeneous datasets. FAVAE~\cite{dehaene2020anomaly} employs VAE to model the distribution of features extracted by the pre-trained model, implicitly simulating richer anomalies and enhancing the model's generalization. Wang~\textit{et al.}~\cite{wang2020image} use VQ-VAE to create a discrete latent space, resample the discrete latent code deviate from the normal distribution, and reconstruct the image using the resampled latent code. VQ-VAE reconstructs images that are closer to the training set's normal images.

\subsubsection{Generative Adversarial Networks}
The stability of the reconstruction model based on Generative Adversarial Networks (GANs) is not as good as that of AE, but the discriminant network has a better effect on some scenes described as follows.

During training, Semantic Context based Anomaly Detection Network (SCADN)~\cite{yan2021learning} masks a portion of the image and reconstructs it with GAN. SCADN detects anomalies for inference by comparing the input image to the reconstruction image. In addition to masking images, AnoSeg~\cite{song2021anoseg} utilizes hard augmentation, adversarial learning, and channel concatenation to generate abnormal samples. AnoSeg then trains GAN to generate normal samples. AnoSeg differs from the AE reconstruction model in that its objective function incorporates both reconstruction loss and adversarial loss. OCR-GAN~\cite{liang2022omni} utilizes the Frequency Decoupling (FD) module to decouple the image into information combinations of different frequencies, and then reconstructs and combines the information of these different frequencies to yield reconstructed images. During inference, the model can identify a statistically significant difference between the frequency distributions of normal and abnormal images.

\subsubsection{Transformer}
Transformer has a higher capacity to represent global information, which gives it the potential to surpass AE and become a new reconstruction network foundation for anomaly detection.
Mishra~\textit{et al.}~\cite{mishra2021vt} propose a transformer-based framework to reconstruct images at the patch level and employ a gaussian mixture density network to localize anomalous regions. You~\textit{et al.}~\cite{you2022adtr} propose ADTR for reconstructing pre-trained features. According to them, the use of transformers prevents well-reconstructed anomalies, making it easy to identify anomalies when reconstruction fails. Lee~\textit{et al.}~\cite{lee2022anovit} introduce a vision transformer-based encoder-decoder model (AnoViT) and assert that AnoViT is superior to the CNN-based $l_2$-CAE in the issue of anomaly detection. HaloAE~\cite{mathian2022haloae} implements transformer into HaloNet~\cite{vaswani2021scaling} and facilitates image reconstruction by reconstructing features to achieve competitive results on the MVTec AD dataset. A common self-supervised learning method for reconstruction-based anomaly detection is the reconstruction of masked images. However, traditional CNNs find it difficult to extract global context information. In order to accomplish this, Pirnay~\textit{et al.}~\cite{pirnay2022inpainting} propose Inpainting Transformer (InTra), which integrates information from larger regions of the input image. InTra is representative of trained-from-scratch methods. Masked Swin Transformer Unet (MSTUnet)~\cite{jiang2022masked} is comparable to InTra, but MSTUnet employs additional enhancements~\cite{perlin1985image} when simulating anomalies, thereby achieving superior results. De~\textit{et al.}~\cite{de2022masked} used the neighbor patch to reconstruct the masked patch and also achieved a powerful reconstruction ability.

\subsubsection{Diffusion Model}
Diffusion model~\cite{ho2020denoising} is a recently popular generative model that can also be utilized for reconstruction-based anomaly detection. AnoDDPM~\cite{wyatt2022anoddpm} is, to the best of our knowledge, the first to apply the diffusion model to industrial image anomaly detection. In comparison to GAN-based methods, AnoDDPM with simplex noise can also capture large anomaly regions without the need for large datasets. When applying the diffusion model to anomaly detection, Teng~\textit{et al.}~\cite{teng2022unsupervised} primarily make two improvements. As a replacement metric for reconstruction loss, a time-dependent gradient value of normal data distribution is used to measure the defects. In addition, they develop a novel T-scales method to reduce the required number of iterations and accelerate the inference process.

\section{Supervised Anomaly Detection}\label{sec:supervised_detection}
Despite the fact that abnormal data is diverse and difficult to collect, it is still possible to collect abnormal samples in real-world scenarios. Therefore, some research focuses on how to train models for anomaly detection using a small number of abnormal samples and a large number of normal samples.

Chu~\textit{et al.}~\cite{chu2020neural} propose a semi-supervised framework for detecting anomalies in the presence of significant data imbalance. They assume that changes in loss values during training can be used to identify abnormal data as features. To achieve this, they train a reinforcement learning-based neural batch sampler to amplify the difference in loss curves between anomalous and non-anomalous regions. FCDD~\cite{liznerski2020explainable} is an unsupervised method that synthesizes abnormal samples for training the OCC model. This concept is transferable to other OCC methods. Venkataramanan~\textit{et al.}~\cite{venkataramanan2020attention} propose a Convolutional Adversarial Variational Autoencoder (CAVGA) with Guided Attention that can be applied equally to cases with and without abnormal images. In an unsupervised setting, CAVGA is guided to focus on all normal regions of an image by an attention expansion loss. CAVGA uses a complementary guided attention loss in the weakly supervised setting to minimize the attention map corresponding to abnormal regions of the image while focusing on normal regions. Bovzivc~\textit{et al.}~\cite{bovzivc2021mixed} examine the influence of image-level supervision information, mixed supervision information, and pixel-level supervision information on surface defect detection tasks within the same deep learning framework. Bovzivc~\textit{et al.} find that a small number of pixel-level annotations can help the model achieve performance comparable to full supervision. DevNet~\cite{pang2021explainable} uses a small number of abnormal samples to realize fine-grained end-to-end differentiable learning. Wan~\textit{et al.}~\cite{wan2022logit} propose a Logit Inducing Loss (LIS) for training with imbalanced data distribution and an Abnormality Capturing Module (ACM) for characterizing anomalous features in order to effectively utilize a small amount of anomalous information. DRA~\cite{ding2022catching}  proposes a framework for learning disentangled representations of seen, pseudo, and latent residual anomalies in order to detect both visible and invisible anomalies.

Besides, a number of studies fail to account for the unbalanced distribution of normal and abnormal samples and rely primarily on abnormal samples for supervised training. Sindagi~\textit{et al.}~\cite{sindagi2017domain} investigate the domain transfer problem of datasets for anomaly detection in various settings. Dual Weighted PCA (DWPCA) is an algorithm proposed by Qiu~\textit{et al.}~\cite{qiu2021effective} for image registration and surface defect detection. An interleaved Deep Artifacts-aware Attention Mechanism (iDAAM) is proposed by Bhattacharya~\textit{et al.}~\cite{bhattacharya2021interleaved} propose to classify multi-object and multi-class defects in abnormal images. Zeng~\textit{et al.}~\cite{zeng2021reference} view anomaly detection as a subset of target detection and designed a Reference-based Defect Detection Network (RDDN) to detect anomalies using template reference and context reference. \reviewer{Song et al.~\cite{song2020saliency} regarded the abnormal part as the salient area of the image, and proposed an effective saliency propagation algorithm for anomaly detection.} Long~\textit{et al.}~\cite{long2021fabric} investigate defect detection in a tactile image, which has obvious benefits for fabric structure defect detection in RGB images. In addition, there are methods that refer to the concept of semantic segmentation. To detect defects in infrared thermal volumetric data, Hu~\textit{et al.}~\cite{hu2020lightweight} propose a hybrid multi-dimensional space and temporal segmentation model. Ferguson~\textit{et al.}~\cite{ferguson2018detection} use Mask Region-based CNN architecture to detect and segment defects in X-ray images simultaneously. There are also numerous modified models on anomaly detection based on the object detection and semantic segmentation model of natural images under full supervision~\cite{Tao2018AutomaticMS, Li2018RealtimeDO, Tabernik2019SegmentationbasedDA}. There are also many weakly supervised object detection methods suitable for anomaly detection~\cite{zhang2022generalized,zhang2020weakly,huang2021scribble}. Here we will not discuss them one by one.


\section{Industrial Manufacturing Setting}\label{sec:manufacturing_setting}

This section introduces the classification standards or application settings that are more appropriate for industrial scenes, namely few-shot anomaly detection, noisy anomaly detection, anomaly synthesis, and 3D anomaly detection.

\subsection{Few-Shot Anomaly Detection}
\reviewer{Few-shot learning is meaningful for data collection and data labeling, which has a great influence on real-world applications. On the one hand, by studying few-shot learning, we can reduce the cost of data collection and data annotation for industrial products. On the other hand, we can solve the problem from the perspective of data and investigate what kind of data is most valuable for industrial image anomaly detection.} Few-Shot Anomaly Detection (FSAD)~\cite{wu2021learning,kamoona2021anomaly} is still in its infancy. There are two settings in FSAD. The first setting is meta-learning~\cite{huang2022registration}. In other words, this setting requires a large amount of images as meta-training dataset. Wu~\textit{et al.}~\cite{wu2021learning} propose a novel architecture, called MetaFormer, that employs meta-learned parameters to achieve high model adaptation capability and instance-aware attention to localize abnormal regions. RegAD~\cite{huang2022registration} trains a model for detecting category-agnostic anomalies. In the test phase, the anomalies are identified by comparing the registered features of the test image and its corresponding normal images. The second setting relies on the vanilla few-shot image learning. PatchCore~\cite{roth2022towards}, SPADE~\cite{cohen2020sub}, PaDim~\cite{defard2021padim} conduct the ablation study on 16 normal training samples. None of them, however, are specialized in few-shot anomaly detection. Hence, it is necessary to develop new algorithms that concentrate on native few-shot anomaly detection tasks. 

\reviewer{Recently, researchers extended the Zero-Shot Anomaly Detection (ZSAD) setting beyond the FSAD setting. The goal of ZSAD is to leverage the generalization power of large models to solve anomaly detection problems without any training, thus completely eliminating the cost of data collection and annotation. MAEDAY~\cite{schwartz2022maeday} uses a pre-trained Masked autoencoder (MAE)~\cite{he2022masked} to tackle the problem. MAEDAY randomly masks parts of an image and restores them using MAE. If the reconstructed region is different from the region before masking, this region is considered as anomalous. WinCLIP~\cite{jeong2023winclip} utilizes another large model called CLIP~\cite{radford2021learning} for ZSAD. Basically, WinCLIP uses the image encoder of CLIP to extract image features. Given the textual descriptions such as ``a photo of a damaged object", WinCLIP uses the text encoder of CLIP to extract the features of these descriptions, and then calculates the similarity between text features and image features. If the similarity is high, the image is ``a photo of a damaged object"; otherwise the image is normal. MAEDAY and WinCLIP demonstrate that zero-shot anomaly detection (ZSAD) is a promising research direction.}

\subsection{Noisy Anomaly Detection}
Noisy learning is a classical problem for anomaly detection. \reviewer{By studying anomaly detection under noisy learning, we can avoid the performance loss caused by labeling errors and reduce false detection in anomaly detection.} Tan~\textit{et al.}~\cite{tan2021trustmae} employ a novel trust region memory update scheme to keep noise feature point away from the memory bank. Yoon~\textit{et al.}~\cite{yoon2021self} use a data refinement approach to improve the robustness of one-class classification model. Qiu~\textit{et al.}~\cite{qiu2022latent} propose a strategy for training an anomaly detector in the presence of unlabeled anomalies, which is compatible with a broad class of models. They create labelled anomalies synthetically and jointly optimize the loss function with normal data and synthesis abnormal data. Chen~\textit{et al.}~\cite{Chen2022DeepOC} introduce an interpolated Gaussian descriptor that learns a one-class Gaussian anomaly classifier trained with adversarially interpolated training samples. However, the majority of the aforementioned approaches have not been verified on real industrial image datasets. In other words, the effectiveness of the existing anomaly detection methods may not be suitable for industrial manufacturing.

\subsection{3D Anomaly Detection}
\reviewer{3D anomaly detection can utilize more spatial information, thereby detecting some information that cannot be contained in RGB images. In some special lighting environments or for some anomalies that are not sensitive to color information, 3D anomaly detection can demonstrate its significant advantages. This research direction is currently receiving significant attention in the academy.} Since the release of MVTec 3D-AD~\cite{bergmann2022beyond} dataset, several papers have focused on anomaly detection in 3D industrial images. Bergmann~\cite{bergmann2022anomaly} introduces a teacher-student model for 3D anomaly detection. The teacher network is trained to acquire general local geometric descriptors by recreating local receptive fields. While the student network is taught to match the local 3D descriptors of the pre-trained teacher network. Horwitz~\textit{et al.}~\cite{horwitz2022back} propose BTF, a method that combines hand-crafted 3D representations (FPFH~\cite{rusu2009fast}) with the representation method of 2D features (PatchCore~\cite{roth2022towards}). Reiss~\textit{et al.}~\cite{reiss2022anomaly} propose that the representational ability of self-supervised learning is temporarily inferior to that of handcrafted features for 3D anomaly detection. Nevertheless, self-supervised characterization still has great potential if large-scale 3D anomaly detection datasets are available. AST~\cite{rudolph2022asymmetric} employs RGB image with depth information to enhance anomaly detection performance. However, most of 3D IAD methods are specialized in RGB-D images, while the 3D dataset in real-world industrial manufacturing consists of point clouds, meaning current 3D IAD methods cannot be directly deployed in industrial manufacturing. Thus, there are still opportunities for 3D IAD advancement.

\subsection{Anomaly Synthesis}
\reviewer{By artificially synthesizing anomalies, we can improve the performance of models with limited data. This research is complementary to the few-shot research. Few-shot learning studies how to improve the model when the data is fixed, and this research studies how to artificially increase the credible data to improve the model performance when the model is fixed. Both of them can reduce the cost of data collection and labeling. There are many unsupervised anomaly detection works that use data augmentation to synthetic anomaly images and significantly improve model performance. For examples, CutPaste~\cite{li2021cutpaste}, DRAEM~\cite{zavrtanik2021draem}, MemSeg~\cite{yang2022memseg} are representative methods.


In addition, some supervised methods use limited abnormal samples to synthesize more abnormal samples for training. Liu et al.~\cite{liu2019multistage} propose a model designed to generate defects on defect-free fabric images for training semantic segmentation. While rippel et al.~\cite{rippel2020gan} use CycleGAN~\cite{zhu2017unpaired} containing ResNet/U-Net as a generator as the basic architecture to transfer defects from one fabric to another. By improving the style transfer network, SDGAN~\cite{niu2020defect} achieves better results than CycleGAN. Wei et al.~\cite{wei2020defective} propose a model named DST to simulate defect samples. First, DST generates a blank mask area on a non-defective image, then DST uses the masked histogram matching module to make the color of the blank mask area consistent with the overall color of the image, and finally DST uses U-NET to perform style transfer to make the generated image more realistic. Wei et al.~\cite{wei2020simulation} propose a model named DSS, which uses conventional GAN to reconstruct defect structures in designated regions of defect-free samples, and then uses DST for style transfer to blend simulated defects into the background. Jain et al.~\cite{jain2020synthetic} try to use DCGAN, ACGCN and InfoGAN to generate defect images by adding noise, which improves the accuracy of classification.  Wang et al.~\cite{wang2021defect} propose DTGAN based on StarGANv2, which adds front-background decoupling and achieves a certain degree of style control and uses the Frechet inception distance (FID~\cite{heusel2017gans}) and kernel inception distance (KID~\cite{binkowski2018demystifying}) to evaluate the quality of image generation. DefectGAN~\cite{zhang2021defect} also believes that defects and normal backgrounds can be layered, and that defects are foreground. DefectGAN generates defect foregrounds and their spatial distribution in the form of style transfer. Although there is a considerable amount of research in this field, unlike other fields that have well-established directions, there is still significant potential for further development.}

\section{Datasets and Metrics}\label{sec:data}

\textbf{Datasets.} Data is a crucial driving factor for machine learning, particularly for deep learning. Principally, the difficulty of getting industrial photos hampers the advancement of image anomaly detection in industrial vision. Table~\ref{tab:datasets_summary} demonstrates that the number and the size of IAD dataset are gradually increasing, but most of them are not generated in a real production line. The promising alternative approach is to fully utilize the industrial simulator to generate anomalous images, possibly reducing the gap between academic research and the demands of industrial manufacturing.

\begin{table}[ht]
\caption{Comparison of datasets for anomaly detection.}
\renewcommand{\arraystretch}{1.2}
\resizebox{\textwidth}{!}{
\begin{tabular}{lllllll}
\hline
 \rowcolor{NavyBlue!10} \textbf{Dataset}      & \textbf{Class} & \textbf{Normal} & \textbf{Abnormal} & \textbf{Total} & \textbf{Annotation Level}  & \textbf{Real or Synthetic} \\ \hline
AITEX~\cite{silvestre2019public}            & 1            & 140              & 105              & 245          & Segmentation mask & Real              \\
BTAD~\cite{mishra2021vt}                    & 3            & - & - & 2,830         & Segmentation mask & Real              \\
DAGM~\cite{wieler2007weakly}                    & 10           & - & - & 11,500        & Segmentation mask & Synthetic         \\
DEEPPCB~\cite{tang2019online}               & 1            & - & - & 1,500         & Bounding box      & Synthetic         \\
Eycandies~\cite{bonfiglioli2022eyecandies}  & 10           & 13,250            & 2,250             & 15,500        & Segmentation mask & Synthetic         \\
Fabric dataset~\cite{tsang2016fabric}       & 1            & 25               & 25               & 50           & Segmentation mask & Synthetic         \\
GDXray~\cite{mery2015gdxray}                & 1            & 0                & 19,407            & 19,407        & Bounding box      & Real              \\
KolektorSDD~\cite{tabernik2020segmentation} & 1            & 347              & 52               & 399          & Segmentation mask & Real              \\
KolektorSDD2~\cite{bovzivc2021mixed}        & 1            & 2,979             & 356              & 3,335         & Segmentation mask & Real              \\
MIAD~\cite{bao2022miad}                     & 7            & 87,500            & 17,500            & 105,000       & Segmentation mask & Synthetic         \\
MPDD~\cite{jezek2021deep}                   & 6            & 1,064             & 282              & 1,346         & Segmentation mask & Real              \\
MTD~\cite{huang2020surface}                 & 1            & - & - & 1,344         & Segmentation mask & Real              \\
MVTec AD~\cite{bergmann2019mvtec}           & 15           & 4,096             & 1,258             & 5,354         & Segmentation mask & Real              \\
MVTec 3D-AD~\cite{bergmann2021mvtec}        & 10           & 2,904             & 948              & 3,852         & Segmentation mask & Real              \\
MVTec LOCO-AD~\cite{bergmann2022beyond}     & 5            & 2,347             & 993              & 3,340         & Segmentation mask & Real              \\
NanoTwice~\cite{carrera2016defect}          & 1            & 5                & 40               & 45           & Segmentation mask & Real              \\
NEU surface defect database~\cite{song2013noise}                & 1            & 0                & 1,800             & 1,800         & Bounding box      & Real              \\
RSDD~\cite{gan2017hierarchical}             & 2            & - & - & 195          & Segmentation mask & Real              \\
Steel Defect Detection~\cite{SDD2019}       & 1            & - & - & 18,076        & Image             & Real              \\
Steel Tube Dataset~\cite{yang2021deep}      & 1            & 0                & 3,408             & 3,408         & Bounding box      & Real              \\
VisA~\cite{zou2022spot}                     & 12           & 9,621             & 1,200             & 10,821        & Segmentation mask & Real             \\ \hline
\end{tabular}}
\label{tab:datasets_summary}
\end{table}

\begin{table}[ht]
\caption{A summary of metrics used for anomaly detection.}
\renewcommand{\arraystretch}{1.2}
\resizebox{\textwidth}{!}{
\begin{tabular}{p{6cm}p{5cm}l}
\hline
 \rowcolor{NavyBlue!10} \textbf{Metric/Level}       & \textbf{Formula}  & \textbf{Remarks/Usage}     \\ \hline
Precision (P)  $\uparrow$ & $P = TP / (TP+FP)$ & True Positive (TP), False Positive (FP) \\
Recall (R)  $\uparrow$ & $R =TP / (TP+FN) $ & False Negative (FN)\\
True Positive Rate (TPR)   $\uparrow$ & $TPR=TP/(TP+FN)$ & \\
False Positive Rate (FPR)  $\downarrow$ & $FPR=FP/FP+TN)$  & True Negative (TN) \\
Area Under the Receiver Operating Characteristic curve (AU-ROC)  $\uparrow$ & $\int_{0}^{1}(TPR)\,{\rm d}(FPR) $   & Classification              \\
Area Under Precision-Recall (AU-PR)  $\uparrow$    & $ \int_{0}^{1}(P)\,{\rm d}(R)$     & Localization, Segmentation \\
Per-Region Overlap (PRO)~\cite{bergmann2021mvtecIJCV}   $\uparrow$      &  $PRO = \frac{1}{N} \sum \limits_i \sum \limits_k \frac{P_i \cap C_{i, k}}{C_{i, k}} $      & \makecell[l]{Total ground truth number (N)/\\ Predicted abnormal pixels (P)/\\ Defect ground truth regions (C)/\\  Segmentation }     \\
Saturated Per-Region Overlap (sPRO)~\cite{bergmann2021mvtec}  $\uparrow$         & $sPRO(P)=\frac{1}{m} \sum \limits_{i=1}^{m} \min(\frac{A_i \cap P}{s_i}, 1)$      & \makecell[l]{Total ground truth number (m)/\\Predicted abnormal pixels (P)/\\ Defect ground truth regions (A)/\\ Corresponding saturation thresholds (s) /\\ Segmentation}         \\
F1 score  $\uparrow$                                                      & $F1=2(P \cdot R) / (P + R)$     &  Classification                \\
Intersection over Union (IoU)~\cite{rahman2016optimizing}   $\uparrow$                                 &  $IoU=(H \cap G)/(H \cup G)$   & \makecell[l]{ Prediction (H), Ground truth (G)/\\ Localization, Segmentation} \\ \hline
\end{tabular}
}
\label{tab:metric-summary}
\end{table}

\textbf{Metrics.}\label{sec:metrics}
Table~\ref{tab:metric-summary} offers a comprehensive review of the metrics in industrial image anomaly detection. The first column denotes the name of the metric and the second column denotes the level. In other words, if the level is up, the larger the metrics value, the better the performance. If the level is down, the lower the metrics value, the better the performance. The third column gives the detail for each metric, especially on how the metric accurately indicates the performance of image anomaly detection. From Table~\ref{tab:metric-summary}, it can be easily observed that most of novel metrics are the variants of natural image segmentation and detection metrics, such as F1 score, AU-ROC or AU-PR. However, these metrics can not correspond to the performance of IAD because the tiny size of anomalies requires a greater weighting than the anomaly-free regions. Hence, the validity of these metrics for IAD remains to be explored. 

\begin{table}[!ht]
\caption{Image AUROC Performance of Different Methods on MVTec AD. The highest and second places are marked in red and blue. \reviewer{All results are reported from the original papers.}}
\renewcommand{\arraystretch}{1.3}
\large
\resizebox{\textwidth}{!}{
\begin{tabular}{l|l|lllllllllllllll|l}
\hline
 \rowcolor{NavyBlue!10} \textbf{Taxonomy} & \textbf{Method}  & \textbf{Bottle} & \textbf{Cable} & \textbf{Capsule} & \textbf{Carpet} & \textbf{Grid} & \textbf{Hazelnut} & \textbf{Leather} & \textbf{Metal Nut} & \textbf{Pill} & \textbf{Screw} & \textbf{Tile} & \textbf{Toothbrush} & \textbf{Transistor} & \textbf{Wood} & \textbf{Zipper} & \textbf{Avg.} \\ \hline
 
\multirow{11}{*}{\textbf{Memory Bank}} & PatchCore~\cite{roth2022towards} & 1.000 & 0.997 & 0.981 & 0.982 & 0.983 & 1.000 & 1.000 & 1.000 & 0.971 & 0.990 & 0.989 & 0.989 & 0.997 & 0.999 & 0.997 & 0.992 \\
 & PatchCore Ensemble~\cite{roth2022towards} & - & - & - & - & - & - & - & - & - & - & - & - & - & - & - & {\color[HTML]{FE0000} \textbf{0.996}} \\
 & CFA~\cite{Lee2022CFACF}  & 1.000 & 0.998 & 0.973 & 0.973 & 0.992 & 1.000 & 1.000 & 1.000 & 0.979 & 0.973 & 0.994 & 1.000 & 1.000 & 0.997 & 0.996 & 0.993 \\
 & FAPM~\cite{kim2022fapm} & 1.000 & 0.995 & 0.986 & 0.993 & 0.980 & 1.000 & 1.000 & 1.000 & 0.960 & 0.952 & 0.994 & 1.000 & 1.000 & 0.993 & 0.995 & 0.990 \\
 & N-pad~\cite{bae2022image} & 1.000 & 0.995 & 0.994 & 0.993 & 0.987 & 1.000 & 1.000 & 1.000 & 0.980 & 0.974 & 1.000 & 1.000 & 0.996 & 0.996 & 0.993 & 0.994 \\
 & N-pad Ensemble~\cite{bae2022image}  & 1.000 & 0.998 & 0.995 & 1.000 & 0.986 & 1.000 & 1.000 & 1.000 & 0.972 & 0.989 & 1.000 & 0.997 & 1.000 & 0.994 & 0.998 & {\color[HTML]{3166FF} \textbf{0.995}} \\
 & MSPB~\cite{tsai2022multi} & 1.000 & 0.988 & 0.972 & 0.934 & 1.000 & 0.996 & 0.993 & 0.978 & 0.977 & 0.941 & 0.962 & 1.000 & 0.989 & 0.997 & 0.995 & 0.981 \\
 & SPD~\cite{zou2022spot} & - & - & - & - & - & - & - & - & - & - & - & - & - & - & - & 0.997 \\
 & SPADE~\cite{cohen2020sub} & - & - & - & - & - & - & - & - & - & - & - & - & - & - & - & 0.855 \\
 & \cite{kim2021semi} & - & - & - & - & - & - & - & - & - & - & - & - & - & - & - & 0.921 \\
 & SOMAD~\cite{li2021anomaly} & 1.000 & 0.988 & 0.988 & 1.000 & 0.939 & 1.000 & 1.000 & 0.997 & 0.986 & 0.955 & 0.987 & 0.986 & 0.945 & 0.992 & 0.977 & 0.979 \\ \hline

\multirow{7}{*}{\textbf{Teacher-Student}} & RD4AD~\cite{deng2022anomaly} & 1.000 & 0.950 & 0.963 & 0.989 & 1.000 & 0.999 & 1.000 & 1.000 & 0.966 & 0.970 & 0.993 & 0.995 & 0.967 & 0.992 & 0.985 & 0.985 \\
 & STFPM~\cite{yamada2021reconstruction} & - & - & - & - & - & - & - & - & - & - & - & - & - & - & - & 0.955 \\
 & Uninformed Students~\cite{bergmann2020uninformed} & 0.918 & 0.865 & 0.916 & 0.695 & 0.819 & 0.937 & 0.819 & 0.895 & 0.935 & 0.928 & 0.912 & 0.863 & 0.701 & 0.725 & 0.933 & 0.857 \\
 & MKD~\cite{salehi2021multiresolution} & 0.994 & 0.892 & 0.805 & 0.793 & 0.780 & 0.984 & 0.951 & 0.736 & 0.827 & 0.833 & 0.916 & 0.922 & 0.856 & 0.943 & 0.932 & 0.877 \\
 & STPM~\cite{Wang2021StudentTeacherFP} & 1.000 & 0.996 & 0.930 & 0.987 & 1.000 & 0.998 & 1.000 & 1.000 & 0.981 & 0.968 & 0.999 & 0.979 & 0.983 & 0.993 & 0.993 & 0.987 \\
 & AST~\cite{rudolph2022asymmetric} & 1.000 & 0.985 & 0.997 & 0.975 & 0.991 & 1.000 & 1.000 & 0.985 & 0.991 & 0.997 & 1.000 & 0.966 & 0.993 & 1.000 & 0.991 & 0.992 \\ \hline
 
\multirow{11}{*}{\textbf{Distribution Map}} &  \cite{rippel2021modeling} & 0.998 & 0.955 & 0.938 & 1.000 & 0.817 & 0.996 & 0.997 & 0.947 & 0.884 & 0.854 & 0.998 & 0.964 & 0.963 & 0.986 & 0.978 & 0.953 \\
 &  \cite{rippel2021transfer} & - & - & - & - & - & - & - & - & - & - & - & - & - & - & - & 0.971 \\
 & PEDENet~\cite{zhang2022pedenet} & - & - & - & - & - & - & - & - & - & - & - & - & - & - & - & 0.928 \\
 & PFM~\cite{wan2022unsupervised} & 1.000 & 0.988 & - & 1.000 & 0.980 & 1.000 & 1.000 & 1.000 & 0.965 & 0.918 & 0.996 & 0.886 & 0.978 & 0.995 & 0.974 & 0.975 \\
 & FYD~\cite{zheng2022focus} & 1.000 & 0.953 & 0.925 & 0.988 & 0.989 & 0.999 & 1.000 & 0.999 & 0.945 & 0.901 & 0.988 & 1.000 & 0.992 & 0.994 & 0.975 & 0.977 \\
 & FastFlow~\cite{yu2021fastflow} & 1.000 & 1.000 & 1.000 & 1.000 & 0.997 & 1.000 & 1.000 & 1.000 & 0.994 & 0.978 & 1.000 & 0.944 & 0.998 & 1.000 & 0.995 & 0.994 \\
 & DifferNet~\cite{rudolph2021same} & 0.990 & 0.959 & 0.869 & 0.929 & 0.840 & 0.993 & 0.971 & 0.961 & 0.888 & 0.963 & 0.994 & 0.986 & 0.911 & 0.998 & 0.951 & 0.949 \\
 & CS-Flow~\cite{rudolph2022fully} & 0.998 & 0.991 & 0.971 & 1.000 & 0.990 & 0.996 & 1.000 & 0.991 & 0.986 & 0.976 & 1.000 & 0.919 & 0.993 & 1.000 & 0.997 & 0.987 \\
 & CFLOW-AD~\cite{gudovskiy2022cflow} & 0.989 & 0.975 & 0.988 & 0.990 & 0.988 & 0.990 & 0.996 & 0.988 & 0.984 & 0.991 & 0.965 & 0.988 & 0.952 & 0.950 & 0.991 & 0.982 \\
 & CS-Flow+AltUB~\cite{kim2022altub} & 1.000 & 0.978 & 0.981 & 0.992 & 1.000 & 1.000 & 1.000 & 0.995 & 0.970 & 0.917 & 0.999 & 0.994 & 0.952 & 0.990 & 0.985 & 0.984 \\ \hline
 
\multirow{10}{*}{\makecell[l]{\textbf{One-Class}\\\textbf{Classfication}}} & Patch SVDD~\cite{yi2020patch} & 0.986 & 0.903 & 0.767 & 0.929 & 0.946 & 0.920 & 0.909 & 0.940 & 0.861 & 0.813 & 0.978 & 1.000 & 0.915 & 0.965 & 0.979 & 0.921 \\
 & SE-SVDD~\cite{hu2021semantic} & 0.986 & 0.977 & 0.985 & 0.989 & 0.972 & 0.980 & 0.987 & 0.983 & 0.967 & 0.986 & 0.923 & 0.993 & 0.972 & 0.951 & 0.979 & 0.975 \\
 & MOCCA~\cite{massoli2021mocca} & 0.950 & 0.760 & 0.820 & 0.860 & 0.870 & 0.800 & 0.980 & 0.850 & 0.820 & 0.840 & 0.890 & 0.970 & 0.880 & 1.000 & 0.840 & 0.875 \\
 & PANDA~\cite{reiss2021panda} & - & - & - & - & - & - & - & - & - & - & - & - & - & - & - & 0.865 \\
 &  \cite{reiss2021mean} & - & - & - & - & - & - & - & - & - & - & - & - & - & - & - & 0.872 \\
 &  \cite{sohn2020learning} & - & - & - & - & - & - & - & - & - & - & - & - & - & - & - & 0.865 \\
 &  \cite{yoa2021self} & 0.918 & 0.883 & 0.965 & 0.894 & 0.881 & 0.962 & 0.985 & 0.926 & 0.964 & 0.972 & 0.919 & 0.958 & 0.883 & 0.892 & 0.954 & 0.930 \\
 & CPC-AD~\cite{de2021contrastive} & 0.998 & 0.880 & 0.641 & 0.809 & 0.983 & 0.996 & 0.990 & 0.845 & 0.921 & 0.897 & 0.957 & 0.878 & 0.925 & 0.803 & 0.993 & 0.901 \\
 & CutPaste~\cite{li2021cutpaste} & 0.982 & 0.812 & 0.982 & 0.939 & 1.000 & 0.983 & 1.000 & 0.999 & 0.949 & 0.887 & 0.946 & 0.994 & 0.961 & 0.991 & 0.999 & 0.961 \\
 & MemSeg~\cite{yang2022memseg} & 1.000 & 0.982 & 1.000 & 0.996 & 1.000 & 1.000 & 1.000 & 1.000 & 0.990 & 0.978 & 1.000 & 1.000 & 0.992 & 0.996 & 1.000 & 0.994 \\ \hline
 
\multirow{21}{*}{\textbf{Reconst.-AE}} & UTAD~\cite{liu2021unsupervised} & - & - & - & - & - & - & - & - & - & - & - & - & - & - & - & 0.900 \\
 & ALT~\cite{yan2021unsupervised} & - & - & - & - & - & - & - & - & - & - & - & - & - & - & - & 0.910 \\
 & \cite{collin2021improved} & 0.980 & 0.890 & 0.740 & 0.890 & 0.970 & 0.940 & 0.890 & 0.730 & 0.840 & 0.740 & 0.990 & 1.000 & 0.910 & 0.950 & 0.940 & 0.890 \\
 & \cite{tao2022unsupervised} & 1.000 & 0.983 & 0.916 & 0.968 & 0.956 & 0.994 & 0.918 & 0.977 & 0.895 & 0.981 & 0.964 & 1.000 & 0.913 & 0.983 & 0.961 & 0.961 \\
 & \cite{hou2021divide} & 0.976 & 0.844 & 0.767 & 0.866 & 0.957 & 0.921 & 0.862 & 0.758 & 0.900 & 0.987 & 0.882 & 0.992 & 0.876 & 0.982 & 0.859 & 0.895 \\
 & EdgRec~\cite{liu2022reconstruction} & 1.000 & 0.979 & 0.955 & 0.974 & 0.997 & 0.984 & 1.000 & 0.973 & 0.990 & 0.899 & 1.000 & 1.000 & 0.998 & 0.940 & 0.983 & 0.978 \\
 & PAE~\cite{kim2022spatial} & 0.999 & 0.948 & 0.956 & 0.989 & 1.000 & 0.981 & 0.973 & 0.965 & 0.975 & 0.956 & 0.985 & 1.000 & 0.990 & 0.987 & 0.991 & 0.980 \\
 & SMAI~\cite{li2020superpixel} & 0.860 & 0.920 & 0.930 & 0.880 & 0.970 & 0.970 & 0.860 & 0.920 & 0.920 & 0.960 & 0.620 & 0.960 & 0.850 & 0.800 & 0.900 & 0.890 \\
 & I3AD~\cite{nakanishi2020iterative} & 0.966 & 0.767 & 0.708 & 0.602 & 0.998 & 0.930 & 0.823 & 0.658 & 0.783 & 0.980 & 0.978 & 0.958 & 0.864 & 0.938 & 0.994 & 0.863 \\
 & \cite{huang2022self} & 0.999 & 0.773 & 0.914 & 0.763 & 1.000 & 0.915 & 0.999 & 0.887 & 0.891 & 0.850 & 0.944 & 1.000 & 0.910 & 0.959 & 0.999 & 0.920 \\
 & RIAD~\cite{zavrtanik2021reconstruction} & 0.999 & 0.819 & 0.884 & 0.842 & 0.996 & 0.833 & 1.000 & 0.885 & 0.838 & 0.845 & 0.987 & 1.000 & 0.909 & 0.930 & 0.981 & 0.917 \\
 & DREAM~\cite{zavrtanik2021draem} & 0.992 & 0.918 & 0.985 & 0.970 & 0.999 & 1.000 & 1.000 & 0.987 & 0.989 & 0.939 & 0.996 & 1.000 & 0.931 & 0.991 & 1.000 & 0.980 \\
 & DSR~\cite{zavrtanik2022dsr} & 1.000 & 0.938 & 0.981 & 1.000 & 1.000 & 0.956 & 1.000 & 0.985 & 0.975 & 0.962 & 1.000 & 0.997 & 0.978 & 0.963 & 1.000 & 0.982 \\
 & NSA~\cite{schluter2022natural} & 0.977 & 0.945 & 0.952 & 0.956 & 0.999 & 0.947 & 0.999 & 0.987 & 0.992 & 0.902 & 1.000 & 1.000 & 0.951 & 0.975 & 0.998 & 0.972 \\
 & \cite{bauer2022self} & 0.950 & 0.960 & 0.980 & 0.990 & 0.990 & 0.980 & 0.990 & 0.950 & 0.980 & 0.990 & 0.970 & 0.980 & 0.970 & 0.970 & 0.990 & 0.980 \\
 & DREAM+SSPCAB~\cite{ristea2022self}  & 0.984 & 0.969 & 0.993 & 0.982 & 1.000 & 1.000 & 1.000 & 1.000 & 0.998 & 0.979 & 1.000 & 1.000 & 0.929 & 0.995 & 1.000 & 0.989 \\
 & DREAM+SSMCTB~\cite{madan2022self}  & 0.994 & 0.941 & 0.971 & 0.968 & 1.000 & 1.000 & 1.000 & 1.000 & 0.988 & 0.990 & 1.000 & 1.000 & 0.960 & 1.000 & 1.000 & 0.987 \\
 & NSA+SSPCAB~\cite{ristea2022self}  & 0.977 & 0.956 & 0.954 & 0.975 & 0.999 & 0.942 & 0.999 & 0.990 & 0.992 & 0.911 & 1.000 & 1.000 & 0.956 & 0.977 & 0.998 & 0.975 \\
 & NSA+SSMCTB~\cite{madan2022self}  & 0.977 & 0.961 & 0.955 & 0.961 & 1.000 & 0.971 & 1.000 & 0.995 & 0.995 & 0.904 & 1.000 & 1.000 & 0.962 & 0.978 & 0.999 & 0.977 \\
 & FAVAE~\cite{dehaene2020anomaly} & 0.999 & 0.950 & 0.804 & 0.671 & 0.970 & 0.993 & 0.675 & 0.852 & 0.821 & 0.837 & 0.805 & 0.958 & 0.932 & 0.948 & 0.972 & 0.879 \\
 &  \cite{wang2020image} & 0.990 & 0.720 & 0.680 & 0.710 & 0.910 & 0.940 & 0.960 & 0.830 & 0.680 & 0.800 & 0.950 & 0.920 & 0.730 & 0.960 & 0.970 & 0.850 \\ \hline
 
\multirow{3}{*}{\textbf{Reconst.-GAN}} & SCADN~\cite{yan2021learning} & 0.957 & 0.856 & 0.765 & 0.504 & 0.983 & 0.833 & 0.659 & 0.624 & 0.814 & 0.831 & 0.792 & 0.981 & 0.863 & 0.968 & 0.846 & 0.818 \\
 & Anoseg~\cite{song2021anoseg} & 0.980 & 0.980 & 0.840 & 0.960 & 0.990 & 0.980 & 0.990 & 0.950 & 0.870 & 0.970 & 0.980 & 0.990 & 0.960 & 0.990 & 0.990 & 0.960 \\
 & OCR-GAN~\cite{liang2022omni} & 0.996 & 0.991 & 0.962 & 0.994 & 0.996 & 0.985 & 0.971 & 0.995 & 0.983 & 1.000 & 0.955 & 0.987 & 0.983 & 0.957 & 0.990 & 0.983 \\ \hline
 
\multirow{6}{*}{\textbf{Reconst.-Transformer}} & ADTR~\cite{you2022adtr} & 1.000 & 0.925 & 0.925 & 1.000 & 0.978 & 0.999 & 1.000 & 0.945 & 0.933 & 0.942 & 1.000 & 0.939 & 0.980 & 0.999 & 0.970 & 0.969 \\
 & AnoViT~\cite{lee2022anovit} & 0.830 & 0.740 & 0.730 & 0.500 & 0.520 & 0.880 & 0.850 & 0.860 & 0.720 & 1.000 & 0.890 & 0.740 & 0.830 & 0.950 & 0.730 & 0.780 \\
 & HaloAE~\cite{mathian2022haloae} & 1.000 & 0.846 & 0.884 & 0.697 & 0.951 & 0.998 & 0.978 & 0.884 & 0.901 & 0.896 & 0.957 & 0.972 & 0.844 & 1.000 & 0.997 & 0.914 \\
 & InTra~\cite{pirnay2022inpainting} & 1.000 & 0.703 & 0.865 & 0.988 & 1.000 & 0.957 & 1.000 & 0.969 & 0.902 & 0.957 & 0.982 & 1.000 & 0.958 & 0.975 & 0.994 & 0.950 \\
 & MSTUnet~\cite{jiang2022masked} & 1.000 & 0.914 & 0.984 & 0.999 & 1.000 & 1.000 & 1.000 & 1.000 & 0.974 & 1.000 & 1.000 & 1.000 & 0.963 & 1.000 & 1.000 & 0.989 \\
 & MeTAL~\cite{de2022masked} & - & - & - & - & - & - & - & - & - & - & - & - & - & - & - & 0.863 \\ \hline

\textbf{Reconst.-Diffusion} & VDD~\cite{teng2022unsupervised} & 1.000 & 0.968 & 0.961 & 0.969 & 1.000 & 0.999 & 0.996 & 0.972 & 0.953 & 0.996 & 0.986 & 0.998 & 0.954 & 0.988 & 0.998 & 0.982 \\ \hline

\multirow{5}{*}{\textbf{Few-Shot}} & Metaformer~\cite{wu2021learning} & 0.991 & 0.971 & 0.875 & 0.940 & 0.859 & 0.994 & - & 0.962 & 0.901 & 0.975 & 0.990 & 1.000 & 0.944 & 0.992 & 0.986 & 0.958 \\
 & RegAD~\cite{huang2022registration} & 0.998 & 0.806 & 0.763 & 0.985 & 0.915 & 0.965 & 1.000 & 0.983 & 0.806 & 0.634 & 0.974 & 0.985 & 0.934 & 0.994 & 0.940 & 0.912 \\
 & \cite{sheynin2021hierarchical} & - & - & - & - & - & - & - & - & - & - & - & - & - & - & - & 0.780 \\
 & RFS~\cite{kamoona2021anomaly} & 1.000 & 0.920 & 0.894 & 0.984 & 0.896 & 0.999 & 1.000 & 0.999 & 0.945 & 0.700 & 0.969 & 0.992 & 0.919 & 0.981 & 0.987 & 0.945 \\
 & MAEDAY~\cite{schwartz2022maeday} & 0.937 & 0.690 & 0.649 & 0.979 & 0.839 & 0.941 & 0.999 & 0.734 & 0.817 & 0.614 & 0.984 & 0.925 & 0.753 & 0.995 & 0.943 & 0.853 \\ \hline
 
\multirow{5}{*}{\textbf{Noisy}} & TrustMAE~\cite{tan2021trustmae} & 0.970 & 0.851 & 0.788 & 0.974 & 0.991 & 0.985 & 0.951 & 0.761 & 0.833 & 0.824 & 0.973 & 0.969 & 0.875 & 0.998 & 0.875 & 0.908 \\
 & \cite{yoon2021self} & - & - & - & - & - & - & - & - & - & - & - & - & - & - & - & 0.937 \\
 & \cite{qiu2022latent} & - & - & - & - & - & - & - & - & - & - & - & - & - & - & - & 0.959 \\
 & IGD~\cite{Chen2022DeepOC} & 1.000 & 0.997 & 0.915 & 0.913 & 0.958 & 0.873 & 0.946 & 0.828 & 0.991 & 0.978 & 0.906 & 0.906 & 0.997 & 0.825 & 0.970 & 0.934 \\
 & SoftPatch~\cite{xisoftpatch} & 0.937 & 0.995 & 0.963 & 0.991 & 0.968 & 1.000 & 1.000 & 0.999 & 0.963 & 0.960 & 0.993 & 0.997 & 0.990 & 0.987 & 0.978 & 0.986 \\ \hline
 
\multirow{3}{*}{\textbf{Supervised}} & CGVGA~\cite{venkataramanan2020attention} & 0.960 & 0.970 & 0.930 & 0.820 & 0.810 & 0.920 & 0.840 & 0.880 & 0.970 & 0.790 & 0.860 & 0.990 & 0.890 & 0.890 & 0.960 & 0.900 \\
 & DevNet~\cite{pang2021explainable} & 0.993 & 0.892 & 0.865 & 0.867 & 0.967 & 1.000 & 0.999 & 0.991 & 0.866 & 0.970 & 0.987 & 0.860 & 0.924 & 0.999 & 0.990 & 0.945 \\
 & DRA~\cite{ding2022catching} & 1.000 & 0.909 & 0.935 & 0.940 & 0.987 & 1.000 & 1.000 & 0.997 & 0.904 & 0.977 & 0.994 & 0.826 & 0.915 & 0.998 & 1.000 & 0.959 \\ \hline
\end{tabular}}
\label{tab:performace_mvtec_imageauc}
\end{table}

\begin{table}[]
\caption{\reviewer{Pixel AUROC and AUPR Performance of Different Methods on MVTec AD. The highest and second places are marked in red and blue. Note that $\ast$ refers to reproduced results by us, while other results are reported from original papers.}}
\resizebox{\textwidth}{!}{
\begin{tabular}{l|l|lllllllllllllll|l}
\hline
\rowcolor{NavyBlue!10} \multicolumn{18}{c}{\textbf{Pixel AU-ROC}}                                                                                                \\  \hline
\rowcolor{NavyBlue!10} \textbf{Taxonomy}                             & \textbf{Methods}                                                       & \textbf{Bottle} & \textbf{Cable} & \textbf{Capsule} & \textbf{Carpet} & \textbf{Grid}  & \textbf{Hazelnut} & \textbf{Leather} & \textbf{Metal\_Nut} & \textbf{Pill}  & \textbf{Screw} & \textbf{Tile}  & \textbf{Toothbrush} & \textbf{Transistor} & \textbf{Wood}  & \textbf{Zipper} & \textbf{Average} \\ \hline
\multirow{8}{*}{Memory Bank}         & PatchCore~\cite{roth2022towards}        & 0.986  & 0.987 & 0.991   & 0.987  & 0.988 & 0.988    & 0.993   & 0.990      & 0.986 & 0.995 & 0.963 & 0.989      & 0.971      & 0.952 & 0.990  & 0.984   \\
                                     & FAPM~\cite{kim2022fapm}                 & 0.982  & 0.985 & 0.990   & 0.989  & 0.978 & 0.986    & 0.990   & 0.982      & 0.980 & 0.990 & 0.980 & 0.987      & 0.982      & 0.940 & 0.986  & 0.980   \\
                                     & N-pad~\cite{bae2022image}               & 0.989  & 0.989 & 0.990   & 0.990  & 0.981 & 0.990    & 0.994   & 0.992      & 0.990 & 0.988 & 0.976 & 0.990      & 0.986      & 0.975 & 0.992  & {\color[HTML]{3166FF} \textbf{0.988}}   \\
                                     & \cite{bae2022image}                          & 0.990  & 0.991 & 0.993   & 0.994  & 0.989 & 0.992    & 0.995   & 0.993      & 0.990 & 0.994 & 0.983 & 0.991      & 0.984      & 0.965 & 0.993  & {\color[HTML]{FE0000} \textbf{0.989}}   \\
                                     & MSPB~\cite{tsai2022multi}               & 0.986  & 0.982 & 0.979   & 0.984  & 0.985 & 0.978    & 0.991   & 0.991      & 0.988 & 0.985 & 0.944 & 0.990      & 0.977      & 0.975 & 0.986  & 0.981   \\
                                     & SPADE~\cite{cohen2020sub}               & 0.984  & 0.972 & 0.990   & 0.975  & 0.937 & 0.991    & 0.976   & 0.981      & 0.965 & 0.989 & 0.874 & 0.979      & 0.941      & 0.885 & 0.965  & 0.960   \\
                                     & SOMAD~\cite{li2021anomaly}              & 0.983  & 0.982 & 0.987   & 0.989  & 0.984 & 0.984    & 0.991   & 0.980      & 0.980 & 0.991 & 0.948 & 0.985      & 0.953      & 0.944 & 0.987  & 0.978   \\
                                     & GCPF~\cite{wan2021industrial}           & 0.975  & 0.957 & 0.977   & 0.990  & 0.978 & 0.981    & 0.993   & 0.959      & 0.970 & 0.975 & 0.961 & 0.973      & 0.907      & 0.951 & 0.982  & 0.969   \\\hline
\multirow{5}{*}{Teacher Student}     & MKD~\cite{salehi2021multiresolution}    & 0.963  & 0.824 & 0.959   & 0.956  & 0.918 & 0.946    & 0.981   & 0.864      & 0.896 & 0.960 & 0.828 & 0.961      & 0.765      & 0.848 & 0.939  & 0.907   \\
                                     & \cite{yamada2021reconstruction}              & 0.989  & 0.976 & 0.989   & 0.990  & 0.993 & 0.991    & 0.990   & 0.986      & 0.971 & 0.994 & 0.968 & 0.990      & 0.881      & 0.964 & 0.985  & 0.977   \\
                                     & \cite{bergmann2020uninformed}                & 0.993  & 0.983 & 0.985   & 0.992  & 0.996 & 0.995    & 0.996   & 0.989      & 0.987 & 0.993 & 0.988 & 0.993      & 0.907      & 0.981 & 0.992  & 0.985   \\
                                     & RD4AD~\cite{deng2022anomaly}            & 0.987  & 0.974 & 0.987   & 0.989  & 0.993 & 0.989    & 0.994   & 0.973      & 0.982 & 0.996 & 0.956 & 0.991      & 0.925      & 0.953 & 0.982  & 0.978   \\
                                     & IKD~\cite{cao2022informative}           & 0.990  & 0.980 & 0.986   & 0.987  & 0.970 & 0.987    & 0.985   & 0.984      & 0.988 & 0.986 & 0.957 & 0.986      & 0.971      & 0.939 & 0.976  & 0.978   \\\hline
\multirow{9}{*}{Distribution Based}  & PEDENet~\cite{zhang2022pedenet}         & 0.984  & 0.971 & 0.943   & 0.922  & 0.959 & 0.970    & 0.976   & 0.973      & 0.960 & 0.972 & 0.926 & 0.979      & 0.982      & 0.900 & 0.962  & 0.959   \\
                                     & PFM\cite{wan2022unsupervised}                & 0.984  & 0.967 & 0.983   & 0.992  & 0.988 & 0.991    & 0.994   & 0.972      & 0.972 & 0.987 & 0.962 & 0.986      & 0.878      & 0.956 & 0.982  & 0.973   \\
                                     & PEFM~\cite{wan2022position}             & 0.985  & 0.983 & 0.985   & 0.992  & 0.992 & 0.992    & 0.994   & 0.970      & 0.970 & 0.990 & 0.966 & 0.992      & 0.984      & 0.965 & 0.986  & 0.983   \\
                                     & FYD~\cite{zheng2022focus}               & 0.983  & 0.975 & 0.986   & 0.985  & 0.968 & 0.987    & 0.992   & 0.982      & 0.973 & 0.987 & 0.968 & 0.989      & 0.981      & 0.996 & 0.982  & 0.982   \\
                                     & FastFlow~\cite{yu2021fastflow}          & 0.977  & 0.984 & 0.991   & 0.994  & 0.983 & 0.991    & 0.995   & 0.985      & 0.992 & 0.994 & 0.963 & 0.989      & 0.973      & 0.970 & 0.987  & 0.985   \\
                                     & CFLOW-AD~\cite{gudovskiy2022cflow}      & 0.990  & 0.976 & 0.990   & 0.993  & 0.990 & 0.989    & 0.997   & 0.986      & 0.990 & 0.989 & 0.980 & 0.989      & 0.980      & 0.967 & 0.991  & 0.986   \\
                                     & CAINNFlow~\cite{yan2022cainnflow}       & 0.985  & 0.987 & 0.989   & 0.994  & 0.989 & 0.993    & 0.997   & 0.991      & 0.985 & 0.997 & 0.975 & 0.996      & 0.976      & 0.955 & 0.987  & 0.986   \\
                                     & CS-Flow+AluUB~\cite{kim2022altub}       & 0.990  & 0.976 & 0.990   & 0.993  & 0.991 & 0.989    & 0.997   & 0.986      & 0.990 & 0.989 & 0.980 & 0.989      & 0.982      & 0.966 & 0.991  & 0.987   \\
                                     & FastFlow+AltUB~\cite{kim2022altub}      & 0.990  & 0.984 & 0.991   & 0.995  & 0.993 & 0.993    & 0.997   & 0.987      & 0.991 & 0.995 & 0.976 & 0.992      & 0.980      & 0.969 & 0.991  & {\color[HTML]{3166FF} \textbf{0.988}}   \\\hline
\multirow{5}{*}{One-Class Classfication}                 & Patch SVDD~\cite{yi2020patch}           & 0.981  & 0.968 & 0.958   & 0.926  & 0.962 & 0.975    & 0.974   & 0.980      & 0.951 & 0.957 & 0.914 & 0.981      & 0.970      & 0.908 & 0.951  & 0.957   \\
                                     & SE-SVDD~\cite{hu2021semantic}           & 0.986  & 0.977 & 0.985   & 0.989  & 0.972 & 0.980    & 0.987   & 0.983      & 0.967 & 0.986 & 0.923 & 0.993      & 0.972      & 0.951 & 0.979  & 0.975   \\
                                     & CPC-AD~\cite{de2021contrastive}         & 0.890  & 0.840 & 0.720   & 0.740  & 0.800 & 0.810    & 0.940   & 0.760      & 0.770 & 0.650 & 0.820 & 0.810      & 0.900      & 0.820 & 0.950  & 0.820   \\
                                     & CutPaste~\cite{li2021cutpaste}          & 0.976  & 0.900 & 0.974   & 0.983  & 0.975 & 0.973    & 0.995   & 0.931      & 0.957 & 0.967 & 0.905 & 0.981      & 0.930      & 0.955 & 0.993  & 0.960   \\
                                     & MemSeg~\cite{yang2022memseg}            & 0.993  & 0.974 & 0.993   & 0.992  & 0.993 & 0.988    & 0.997   & 0.993      & 0.995 & 0.980 & 0.995 & 0.994      & 0.973      & 0.980 & 0.988  & {\color[HTML]{3166FF} \textbf{0.988}}   \\\hline
\multirow{20}{*}{Recons.-AE}         & \cite{chung2020unsupervised}                 & 0.940  & 0.840 & 0.920   & 0.950  & 0.970 & 0.970    & 0.850   & 0.890      & 0.940 & 0.970 & 0.810 & 0.980      & 0.880      & 0.790 & 0.870  & 0.920   \\
                                     & DFR~\cite{yang2020dfr}                  & 0.970  & 0.920 & 0.990   & 0.970  & 0.980 & 0.990    & 0.980   & 0.930      & 0.970 & 0.990 & 0.870 & 0.990      & 0.800      & 0.930 & 0.960  & 0.950   \\
                                     & ALT~\cite{yan2021unsupervised}          & 0.964  & 0.908 & 0.988   & 0.971  & 0.995 & 0.991    & 0.989   & 0.976      & 0.985 & 0.993 & 0.955 & 0.977      & 0.914      & 0.962 & 0.975  & 0.969   \\
                                     & AESc-Stain~\cite{collin2021improved}    & 0.880  & 0.840 & 0.930   & 0.910  & 0.950 & 0.890    & 0.870   & 0.620      & 0.850 & 0.950 & 0.790 & 0.930      & 0.780      & 0.840 & 0.900  & 0.860   \\
                                     & \cite{tao2022unsupervised}                   & 0.964  & 0.971 & 0.983   & 0.991  & 0.981 & 0.988    & 0.992   & 0.983      & 0.967 & 0.993 & 0.909 & 0.986      & 0.870      & 0.941 & 0.982  & 0.967   \\
                                     & EdgRec~\cite{liu2022reconstruction}     & 0.983  & 0.977 & 0.952   & 0.994  & 0.992 & 0.994    & 0.997   & 0.980      & 0.987 & 0.977 & 0.986 & 0.992      & 0.943      & 0.914 & 0.987  & 0.977   \\
                                     & \cite{kim2022spatial}                        & 0.974  & 0.975 & 0.961   & 0.993  & 0.993 & 0.985    & 0.988   & 0.962      & 0.967 & 0.997 & 0.983 & 0.977      & 0.981      & 0.931 & 0.986  & 0.977   \\
                                     & I3AD~\cite{nakanishi2020iterative}      & 0.950  & 0.795 & 0.854   & 0.850  & 0.987 & 0.756    & 0.938   & 0.526      & 0.725 & 0.959 & 0.788 & 0.969      & 0.651      & 0.776 & 0.962  & 0.832   \\
                                     & ~\cite{huang2022self}                   & 0.959  & 0.821 & 0.984   & 0.944  & 0.990 & 0.974    & 0.996   & 0.896      & 0.978 & 0.989 & 0.902 & 0.989      & 0.801      & 0.869 & 0.990  & 0.939   \\
                                     & RIAD~\cite{zavrtanik2021reconstruction} & 0.984  & 0.842 & 0.928   & 0.963  & 0.988 & 0.961    & 0.994   & 0.925      & 0.957 & 0.988 & 0.891 & 0.989      & 0.877      & 0.858 & 0.978  & 0.942   \\
                                     & DREAM~\cite{zavrtanik2021draem}         & 0.991  & 0.947 & 0.943   & 0.955  & 0.997 & 0.997    & 0.986   & 0.995      & 0.976 & 0.976 & 0.992 & 0.981      & 0.909      & 0.964 & 0.988  & 0.973   \\
                                     & NSA~\cite{schluter2022natural}          & 0.983  & 0.960 & 0.976   & 0.955  & 0.992 & 0.976    & 0.995   & 0.984      & 0.985 & 0.965 & 0.993 & 0.949      & 0.880      & 0.907 & 0.942  & 0.968   \\
                                     & DREAM+SSPCAB~\cite{ristea2022self}      & 0.988  & 0.960 & 0.931   & 0.950  & 0.995 & 0.998    & 0.995   & 0.989      & 0.975 & 0.998 & 0.993 & 0.981      & 0.870      & 0.968 & 0.990  & 0.972   \\
                                     & DREAM+SSMCTB~\cite{madan2022self}       & 0.992  & 0.955 & 0.934   & 0.958  & 0.997 & 0.995    & 0.976   & 0.993      & 0.974 & 0.995 & 0.993 & 0.990      & 0.891      & 0.948 & 0.990  & 0.972   \\
                                     & NSA+SSPCAB~\cite{ristea2022self}        & 0.983  & 0.966 & 0.972   & 0.975  & 0.992 & 0.979    & 0.995   & 0.979      & 0.988 & 0.962 & 0.992 & 0.953      & 0.871      & 0.904 & 0.945  & 0.964   \\
                                     & NSA+SSMCTB~\cite{madan2022self}         & 0.984  & 0.975 & 0.979   & 0.956  & 0.992 & 0.979    & 0.996   & 0.983      & 0.984 & 0.964 & 0.991 & 0.954      & 0.883      & 0.935 & 0.947  & 0.967   \\
                                     & \cite{dehaene2019iterative}                  & 0.922  & 0.910 & 0.917   & 0.735  & 0.961 & 0.976    & 0.925   & 0.907      & 0.930 & 0.945 & 0.654 & 0.985      & 0.919      & 0.838 & 0.869  & 0.893   \\
                                     & \cite{liu2020towards}                        & 0.870  & 0.900 & 0.740   & 0.780  & 0.730 & 0.980    & 0.950   & 0.940      & 0.830 & 0.970 & 0.800 & 0.940      & 0.930      & 0.770 & 0.780  & 0.860   \\
                                     & FAVAE~\cite{dehaene2020anomaly}         & 0.963  & 0.969 & 0.976   & 0.960  & 0.993 & 0.987    & 0.981   & 0.966      & 0.953 & 0.993 & 0.714 & 0.987      & 0.984      & 0.899 & 0.968  & 0.953   \\
                                     & \cite{wang2020image}                         & 0.950  & 0.950 & 0.930   & 0.940  & 0.990 & 0.950    & 0.990   & 0.910      & 0.950 & 0.960 & 0.880 & 0.970      & 0.910      & 0.870 & 0.980  & 0.940   \\\hline
\multirow{2}{*}{Recons.-GAN}         & SCADN~\cite{yan2021learning}            & 0.696  & 0.814 & 0.687   & 0.649  & 0.796 & 0.884    & 0.763   & 0.754      & 0.747 & 0.876 & 0.677 & 0.901      & 0.689      & 0.672 & 0.670  & 0.752   \\
                                     & Anoseg~\cite{song2021anoseg}            & 0.990  & 0.990 & 0.900   & 0.990  & 0.990 & 0.990    & 0.980   & 0.990      & 0.940 & 0.910 & 0.980 & 0.960      & 0.960      & 0.980 & 0.980  & 0.970   \\\hline
\multirow{6}{*}{Recons.-Transformer} & ADTR~\cite{you2022adtr}                 & 0.980  & 0.970 & 0.991   & 0.988  & 0.942 & 0.988    & 0.986   & 0.968      & 0.987 & 0.993 & 0.959 & 0.992      & 0.978      & 0.930 & 0.976  & 0.975   \\
                                     & AnoViT~\cite{lee2022anovit}             & 0.860  & 0.890 & 0.910   & 0.650  & 0.830 & 0.940    & 0.890   & 0.880      & 0.860 & 0.920 & 0.570 & 0.900      & 0.800      & 0.850 & 0.760  & 0.830   \\
                                     & HaloAE~\cite{mathian2022haloae}         & 0.919  & 0.876 & 0.978   & 0.894  & 0.831 & 0.978    & 0.985   & 0.852      & 0.915 & 0.990 & 0.785 & 0.929      & 0.875      & 0.911 & 0.960  & 0.912   \\
                                     & InTra~\cite{pirnay2022inpainting}       & 0.971  & 0.910 & 0.977   & 0.992  & 0.988 & 0.983    & 0.995   & 0.933      & 0.983 & 0.995 & 0.944 & 0.989      & 0.961      & 0.887 & 0.992  & 0.966   \\
                                     & MSTUnet~\cite{jiang2022masked}          & 0.990  & 0.899 & 0.957   & 0.983  & 0.997 & 0.993    & 0.995   & 0.993      & 0.976 & 0.974 & 0.997 & 0.991      & 0.746      & 0.980 & 0.989  & 0.964   \\
                                     & \cite{de2022masked}                          & 0.850  & 0.701 & 0.891   & 0.712  & 0.884 & 0.959    & 0.976   & 0.773      & 0.852 & 0.901 & 0.771 & 0.975      & 0.860      & 0.836 & 0.750  & 0.846   \\\hline
Recons.-Diffusion Model              & VDD~\cite{teng2022unsupervised}         & 0.979  & 0.975 & 0.986   & 0.989  & 0.997 & 0.992    & 0.993   & 0.979      & 0.960 & 0.996 & 0.944 & 0.983      & 0.952      & 0.951 & 0.993  & 0.978   \\\hline
\multirow{2}{*}{Few-Shot}            & Metaformer~\cite{wu2021learning}        & 0.888  & 0.937 & 0.879   & 0.878  & 0.865 & 0.886    & 0.959   & 0.869      & 0.930 & 0.954 & 0.881 & 0.877      & 0.926      & 0.848 & 0.936  & 0.901   \\
                                     & MAEDAY~\cite{schwartz2022maeday}        & 0.959  & 0.842 & 0.953   & 0.982  & 0.966 & 0.983    & 0.994   & 0.684      & 0.913 & 0.974 & 0.901 & 0.922      & 0.860      & 0.929 & 0.962  & 0.922   \\\hline
\multirow{2}{*}{Noisy}               & TrustMAE~\cite{tan2021trustmae}         & 0.934  & 0.929 & 0.874   & 0.985  & 0.975 & 0.985    & 0.981   & 0.918      & 0.899 & 0.976 & 0.825 & 0.981      & 0.927      & 0.926 & 0.978  & 0.939   \\
                                     & SoftPatch~\cite{xisoftpatch}            & 0.975  & 0.971 & 0.989   & 0.989  & 0.974 & 0.924    & 0.993   & 0.983      & 0.976 & 0.969 & 0.954 & 0.985      & 0.936      & 0.929 & 0.986  & 0.969   \\\hline
\multirow{2}{*}{Supervised AD}       & FCDD~\cite{liznerski2020explainable}    & 0.960  & 0.930 & 0.950   & 0.990  & 0.950 & 0.970    & 0.990   & 0.980      & 0.970 & 0.930 & 0.980 & 0.950      & 0.900      & 0.940 & 0.980  & 0.960   \\
                                     & DevNet~\cite{pang2021explainable}       & 0.951  & 0.920 & 0.938   & 0.963  & 0.935 & 0.959    & 0.990   & 0.876      & 0.859 & 0.897 & 0.950 & 0.819      & 0.839      & 0.900 & 0.973  & 0.918   \\ \hline
\rowcolor{NavyBlue!10} \multicolumn{18}{c}{\textbf{Pixel AU-PR}}                                                       \\ \hline
\rowcolor{NavyBlue!10} Taxonomy                             & Methods                                                       & Bottle & cable & capsule & carpet & grid  & hazelnut & leather & metal\_nut & pill  & screw & tile  & toothbrush & transistor & wood  & zipper & average \\ \hline
\multirow{4}{*}{Memory Bank}     & CFA$\ast$~\cite{xie2023iad}      & 0.726 & 0.691 & 0.394 & 0.484 & 0.721 & 0.406 & 0.580 & 0.440 & 0.525 & 0.317 & 0.584 & 0.405 & 0.802 & 0.764 & 0.237 & 0.538 \\
                                 & PatchCore$\ast$~\cite{xie2023iad} & 0.768 & 0.653 & 0.442 & 0.627 & 0.325 & 0.537 & 0.456 & 0.870 & 0.777 & 0.354 & 0.546 & 0.372 & 0.610 & 0.477 & 0.595 & 0.561 \\
                                 & SPADE$\ast$~\cite{xie2023iad}     & 0.699 & 0.234 & 0.265 & 0.573 & 0.321 & 0.418 & 0.531 & 0.438 & 0.518 & 0.236 & 0.662 & 0.482 & 0.793 & 0.452 & 0.437 & 0.471 \\
                                 & PaDiM$\ast$~\cite{xie2023iad}     & 0.730 & 0.343 & 0.334 & 0.497 & 0.580 & 0.374 & 0.452 & 0.394 & 0.602 & 0.249 & 0.517 & 0.406 & 0.713 & 0.423 & 0.166 & 0.452 \\ \hline
Distribution Based               & FastFlow$\ast$~\cite{xie2023iad}  & 0.688 & 0.276 & 0.418 & 0.205 & 0.539 & 0.455 & 0.351 & 0.276 & 0.633 & 0.311 & 0.259 & 0.525 & 0.609 & 0.378 & 0.045 & 0.398 \\ \hline
\multirow{2}{*}{Teacher Student} & RD4AD$\ast$~\cite{xie2023iad}     & 0.787 & 0.527 & 0.451 & 0.574 & 0.492 & 0.621 & 0.482 & 0.791 & 0.784 & 0.536 & 0.532 & 0.518 & 0.549 & 0.482 & 0.571 & 0.580 \\
                                 & STPM$\ast$~\cite{xie2023iad}      & 0.732 & 0.327 & 0.706 & 0.470 & 0.299 & 0.400 & 0.546 & 0.363 & 0.669 & 0.452 & 0.596 & 0.491 & 0.701 & 0.801 & 0.220 & 0.518  \\ \hline
\multirow{5}{*}{Recons.-AE}          & EdgRec~\cite{liu2022reconstruction}     & 0.779  & 0.708 & 0.405   & 0.798  & 0.466 & 0.817    & 0.664   & 0.799      & 0.793 & 0.429 & 0.845 & 0.592      & 0.712      & 0.548 & 0.525  & 0.658   \\
                                     & DREAM~\cite{zavrtanik2021draem}         & 0.865  & 0.524 & 0.494   & 0.535  & 0.657 & 0.929    & 0.753   & 0.963      & 0.485 & 0.582 & 0.923 & 0.447      & 0.507      & 0.777 & 0.815  & 0.684   \\
                                     & DSR~\cite{zavrtanik2022dsr}             & 0.915  & 0.704 & 0.533   & 0.782  & 0.680 & 0.873    & 0.625   & 0.675      & 0.657 & 0.525 & 0.939 & 0.742      & 0.411      & 0.684 & 0.785  & 0.702   \\
                                     & DREAM+SSPCAB~\cite{ristea2022self}      & 0.879  & 0.572 & 0.502   & 0.594  & 0.611 & 0.926    & 0.760   & 0.981      & 0.524 & 0.720 & 0.950 & 0.510      & 0.480      & 0.771 & 0.771  & {\color[HTML]{3166FF} \textbf{0.703}}   \\
                                     & DREAM+SSMCTB~\cite{madan2022self}       & 0.899  & 0.616 & 0.520   & 0.552  & 0.697 & 0.891    & 0.655   & 0.947      & 0.469 & 0.701 & 0.957 & 0.690      & 0.458      & 0.756 & 0.765  & {\color[HTML]{FE0000} \textbf{0.705}}   \\ \hline 
\end{tabular}}
\label{tab:performace_mvtec_pixel}
\end{table}

\begin{figure}[h]
\centering
    \includegraphics[width=1\linewidth]{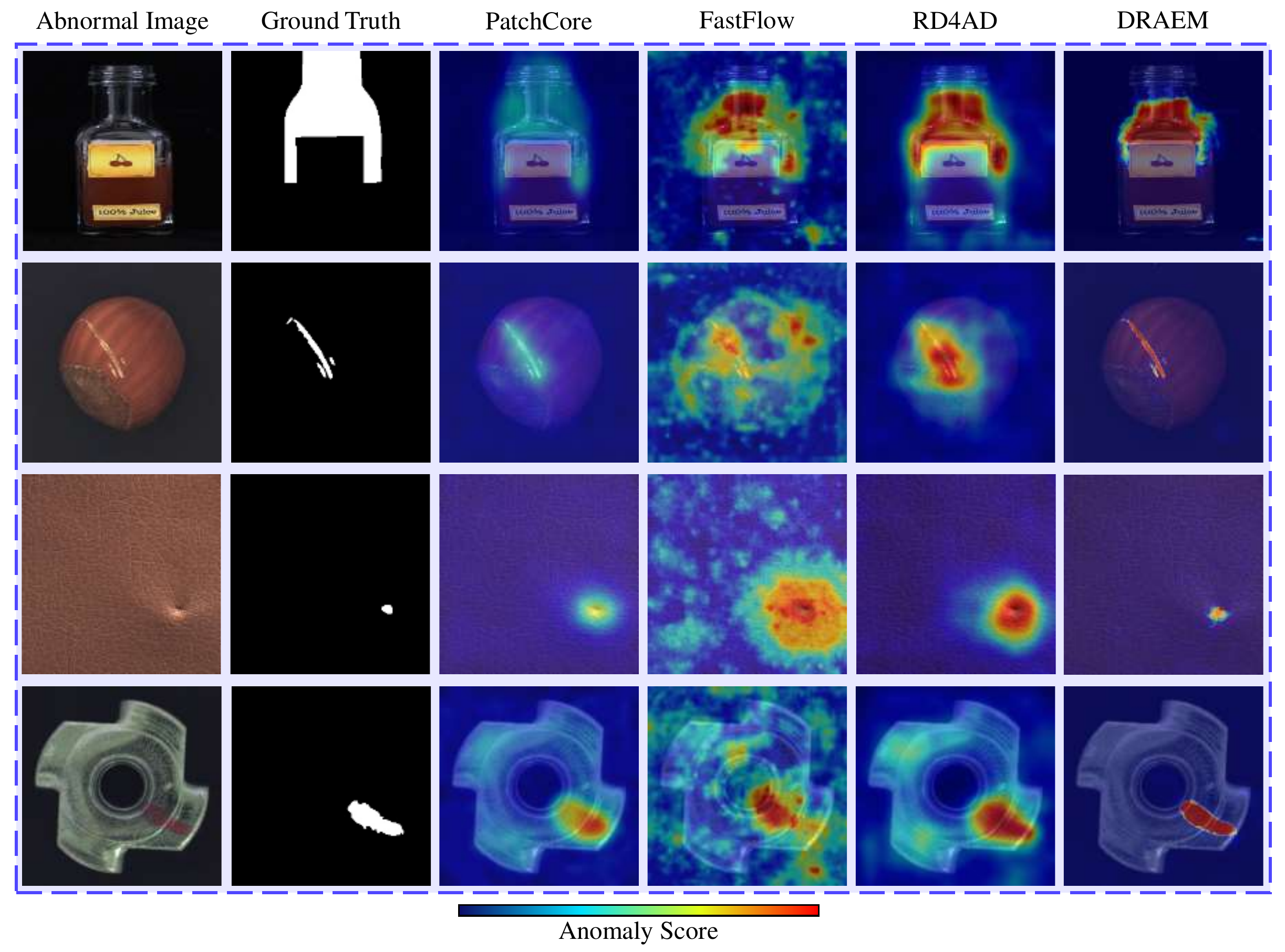}
    \caption{\reviewer{Visualization of results from representative methods. Note that the visualization results are from the open-source code reproduction.}}
    \label{fig:IAD_Survey_VIS}
\end{figure}

\section{Total Performance Analysis}\label{sec:total_performance}

\reviewer{Table~\ref{tab:performace_mvtec_imageauc} and Table~\ref{tab:performace_mvtec_pixel} show the statistical result of current IAD performance on MVTec AD. Fig.~\ref{fig:IAD_Survey_VIS} supports the results of Table~\ref{tab:performace_mvtec_imageauc}: even if different methods have similar performance in image classification, there are still significant differences in pixel-level segmentation.} We provide a deep analysis of the performance of current IAD methods and unlock meaningful insights as below:
\xgy{\begin{itemize}
    \item Regarding the identification of image-level anomaly detection tasks, memory bank-based approaches are the most effective neural network design. However, they are inadequate at detecting pixel-level anomalies.  
    \item Ensemble learning can dramatically improve the performance of state-of-the-art anomaly detection methods.
    \item SSPCAB~\cite{ristea2022self} can be seamlessly integrated into cutting-edge methods and significantly enhance the performance of reconstruction-based methods.
    \item The gap between few-shot IAD and vanilla IAD is narrowing. In other words, we may utilize data distillation algorithms to lower the amount of the dataset used for anomaly detection. 
    \item \reviewer{Without using ensemble learning, MemSeg~\cite{yang2022memseg} achieved the SOTA result on image-level anomaly classification, which is mainly due to the use of the U-Net~\cite{ronneberger2015u} framework. DRAEM~\cite{zavrtanik2021draem} also uses U-Net to outperform other methods on pixel-level anomaly segmentation. The effectiveness of MemSeg and DRAEM demonstrates the superiority of the segmentation module in anomaly detection. Artificial supervision is usually inferior to real supervision, and segmentation models trained with artificial supervision often perform worse. However, even when using artificially generated anomalies as supervisory information, these methods with segmentation modules still outperform other methods without segmentation modules on classification and segmentation tasks. We can conclude that the segmentation module is beneficial for anomaly detection tasks.}
    \item \reviewer{AU-PR is more valuable than AU-ROC for segmentation tasks~\cite{zou2022spot}. As shown in Table~\ref{tab:performace_mvtec_pixel}, reconstruction-based methods outperform other methods on the pixel AU-PR metric. As for Fig.~\ref{fig:IAD_Survey_VIS}, the detection result of DREAM is closest to the ground truth. It results in sharper edges and fewer false detection regions. We can infer from statistical data and visualizations that reconstruction-based methods are more suitable for segmentation tasks.}
\end{itemize}}


\section{Future Directions}\label{sec:future_direction}
We outline several intriguing future directions as follows:
\begin{itemize}
    \item We should build up a multi-modalities IAD Dataset. In actual assembly lines, RGB images are insufficient to detect anomalies. Hence, we may employ additional modalities information, such as X-ray and ultrasound, to enhance anomaly detection performance.
    \item Given that test samples are sequentially streamed on the product line, most IAD methods are incapable of making instantaneous predictions upon the arrival of a new test sample. In industrial manufacturing, the inference speed of IAD should be addressed in addition to its accuracy. Adopting multi-objective evolutionary neural architecture search algorithms to find the optimal trade-off architecture is thus a promising approach.
    \item The majority of IAD methods use ImageNet pre-trained models to extract the features from industrial images, which inevitably results in the feature drift issue. Consequently, there is a pressing need to construct a pre-trained model for industrial images.
    \item \reviewer{Most anomaly detection methods focus on the unsupervised setting. Although this setting can reduce the cost of data labeling, it greatly curbs the development of segmentation-based methods. Unsupervised methods and supervised methods should complement each other, and the main reason for the slow development of supervised methods in recent years is the lack of a large number of labeled data sets. Therefore, it is necessary to propose a fully supervised anomaly detection dataset with pixel-level annotations in the future.}
    \item Previously, we focused on developing data augmentation method for normal images. However, we have not made much effort on synthesizing abnormal samples via data augmentation. In industrial manufacturing, it is very difficult to collect a large number of abnormal samples since most of the production lines are faultless. Hence, more attention should be paid to abnormal synthesis methods in the future, like CutPaste~\cite{li2021cutpaste}, DRAEM~\cite{zavrtanik2021draem} and MemSeg~\cite{yang2022memseg}.
    \item \reviewer{Current anomaly detection algorithms often focus on detection accuracy, while ignoring the storage size and efficiency of the models. This leads to high computation costs and limits the application of anomaly detection to the production end of enterprises. Therefore, it is necessary to design lightweight but efficient anomaly detection models.}
    \item \reviewer{Currently, image anomaly detection algorithms can be mainly categorized into two tasks: industrial image anomaly detection and medical image anomaly detection. Although medical images have more modalities than industrial images~\cite{zhang2020exploring, zhang2021cross,fang2023reliable}, the two tasks share many similarities in terms of data and experimental settings. However, few studies have explored how to unify these two tasks. One reason for this is the domain differences between medical and industrial image datasets, and another reason is the lack of a good baseline and benchmark for comparison. It would be very meaningful to establish a unified framework for both industrial and medical image anomaly detection at the data or method level.}
\end{itemize}

\section{Conclusions}
\xgy{In this paper, we provide a literature review on image anomaly detection in industrial manufacturing, focusing on the level of supervision, the design of neural network architecture, the types and properties of datasets and the evaluation metrics. \reviewer{In particular, we characterize the promising setting from industrial manufacturing and review current IAD algorithms in our proposed setting.} In addition, we investigate in depth which network architecture design can considerably improve anomaly detection performance. In the end, we highlight several exciting future research directions for image anomaly detection.}


\section*{Acknowledgments} 
This work was partly supported by the National Key R\&D Program of China (Grant NO. 2022YFF1202903) and the National Natural Science Foundation of China (Grant NO. 62122035 and 62206122). Y. Jin is funded by an Alexander von Humboldt Professorship for Artificial Intelligence endowed by the Federal Ministry of Education and Research of Germany.

\bibliography{reference}

\clearpage




\begin{figure}[h]%
\centering
\includegraphics[width=0.3\textwidth]{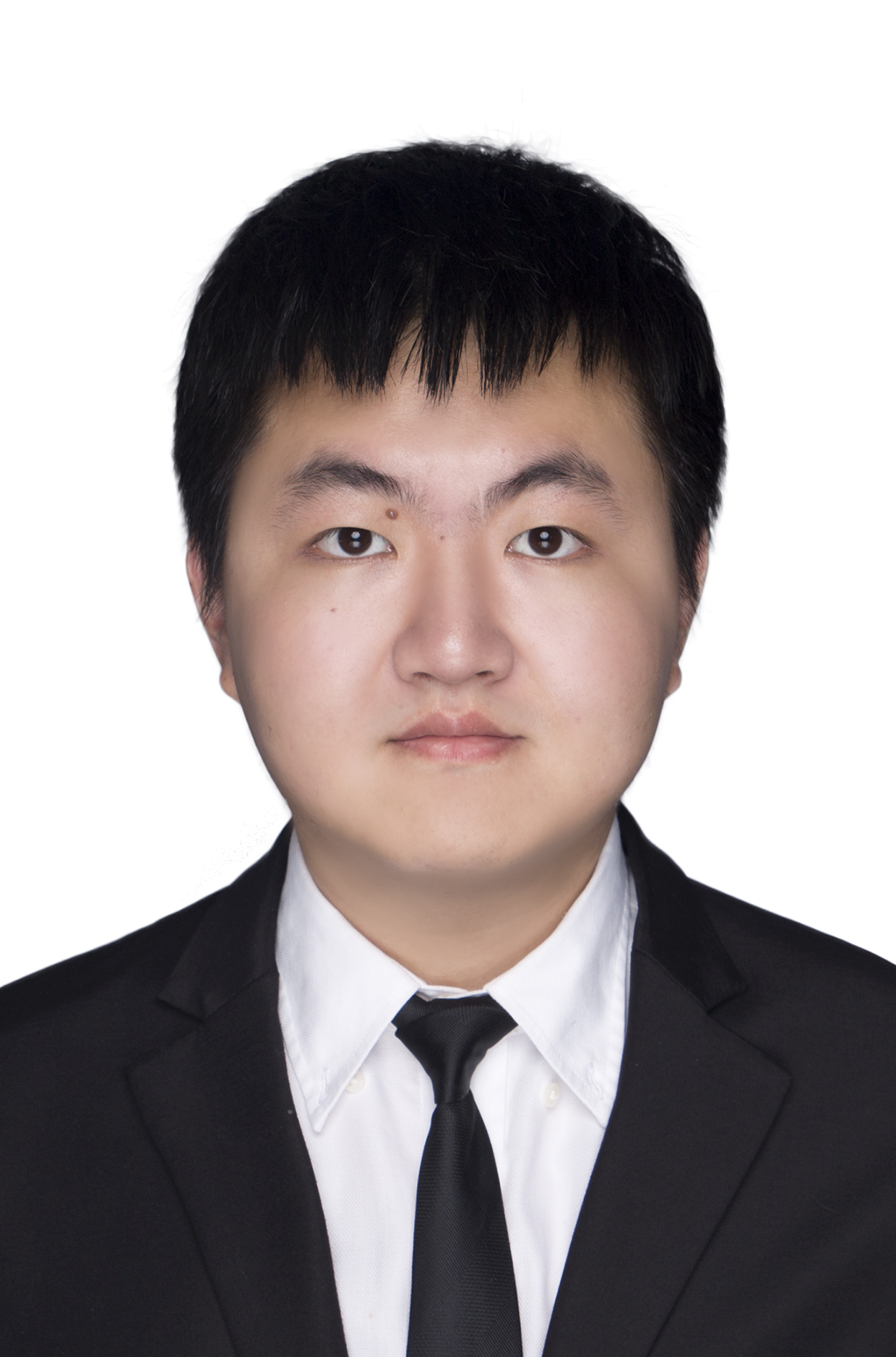}
\end{figure}
\noindent{\bf Jiaqi Liu}\quad received his BS degree from Dalian University of Technology in 2019. He is pursuing the MS degree from Southern University of Science and Technology under the supervision of Professor Feng Zheng. His research interest focuses on image anomaly detection.

E-mail: liujq32021@mail.sustech.edu.cn (Contributed equally)

ORCID iD: 0000-0002-2153-8411

\vspace{1em}

\begin{figure}[h]%
\centering
\includegraphics[width=0.3\textwidth]{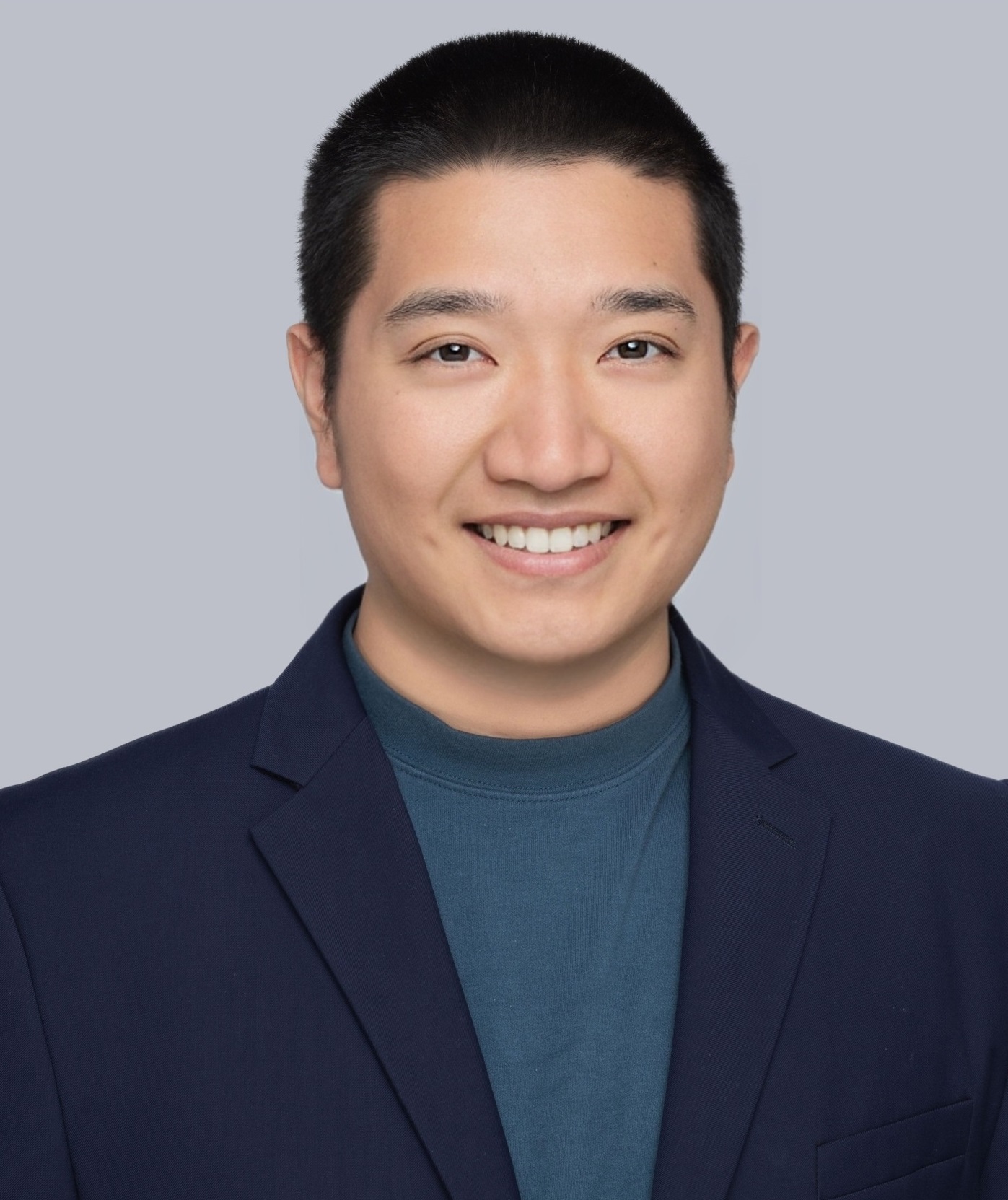}
\end{figure}
\noindent{\bf Guoyang Xie}\quad received the Bachelor and MPhil Degrees from University of Electronic Science and Technology of China, Hong Kong University of Science and Technology in 2009 and 2013, respectively. He is pursuing the PhD degree from University of Surrey. Prior to that, he was the Principle Perception Algorithm Engineer in Baidu and GAC, respectively. His research interests include anomaly detection, medical imaging, neural architecture search and federated learning.

E-mail: guoyang.xie@surrey.ac.uk (Contributed equally)

ORCID iD: 0000-0001-8433-8153

\vspace{1em}

\begin{figure}[h]%
\centering
\includegraphics[width=0.3\textwidth]{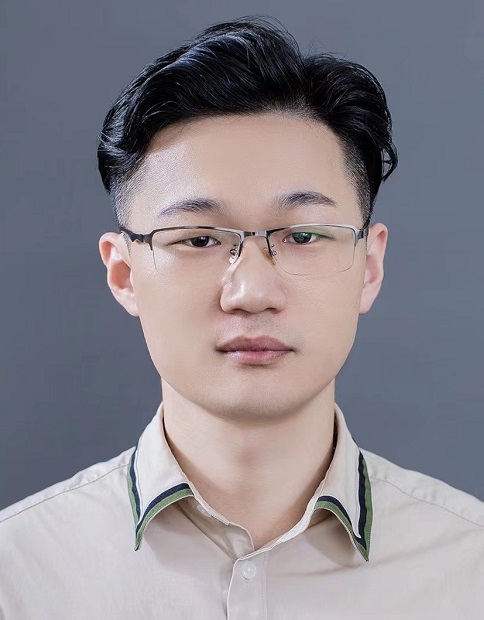}
\end{figure}
\noindent{\bf Jinbao Wang}\quad received the Ph.D. degree from the University of Chinese Academy of Sciences (UCAS) in 2019. He is currently a Research Assistant Professor with the Southern University of Science and Technology (SUSTech), Shenzhen, China. His research interests include machine learning, computer vision, image anomaly detection, and graph representation learning.

E-mail: linkingring@163.com (Contributed equally)

ORCID iD: 0000-0001-5916-8965

\vspace{1em}

\begin{figure}[h]%
\centering
\includegraphics[width=0.3\textwidth]{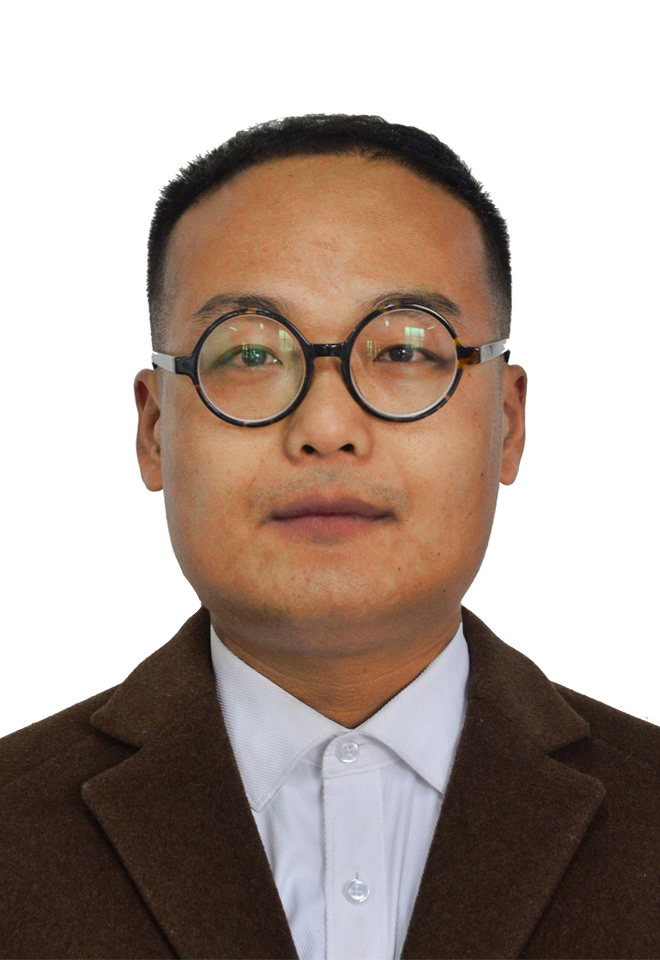}
\end{figure}
\noindent{\bf Shangnian Li}\quad received the Bachelor and Master Degrees from Huaiyin Institute Of Technology, Beijing Union University
in 2012 and 2016, respectively. He is currently the research assistant of Sustech VIP lab.
Prior to that, he was the vehicle networking engineer in Beijing Xiangzhi Technology Co., Ltd. His research interests include internet of things, anomaly detection.

E-mail: lisn3@mail.sustech.edu.cn

\vspace{1em}

\begin{figure}[h]%
\centering
\includegraphics[width=0.3\textwidth]{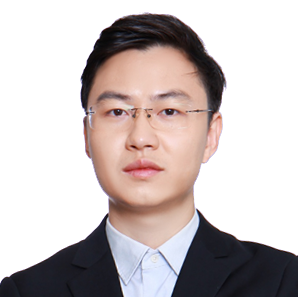}
\end{figure}
\noindent{\bf Chengjie Wang}\quad received the BS degree in computer science from Shanghai Jiao
Tong University, Shanghai, China, in 2011, and double MS degrees in computer science from Shanghai Jiao Tong University, and Waseda University, Tokyo, Japan, in
2014. He is currently the Research Director of YouTu Lab, Tencent. And he is pursuing his Ph.D. degree at Department of Computer Science and Engineering, Shanghai
Jiao Tong University, China. He has authored or coauthored more than 90 refereed
papers on major Computer Vision and Artificial Intelligence Conferences, such as
CVPR, ICCV, ECCV, AAAI, IJCAI, and NeurIPS, and holds more than 100 patents in his
research areas. His research interests include computer vision and machine learning.

E-mail: jasoncjwang@tencent.com

\vspace{1em}

\begin{figure}[h]%
\centering
\includegraphics[width=0.3\textwidth]{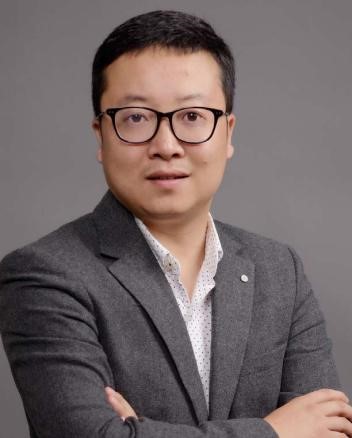}
\end{figure}
\noindent{\bf Feng Zheng}\quad received the Ph.D. degree from The University of Sheffield, Sheffield, U.K., in 2017. He is currently an Assistant Professor with the Department of Computer Science and Engineering, Southern University of Science and Technology, Shenzhen, China. His research interests include machine learning, computer vision, and human-computer interaction.

E-mail: f.zheng@ieee.com (Corresponding author)

ORCID iD: 0000-0002-1701-9141

\vspace{1em}

\begin{figure}[h]%
\centering
\includegraphics[width=0.3\textwidth]{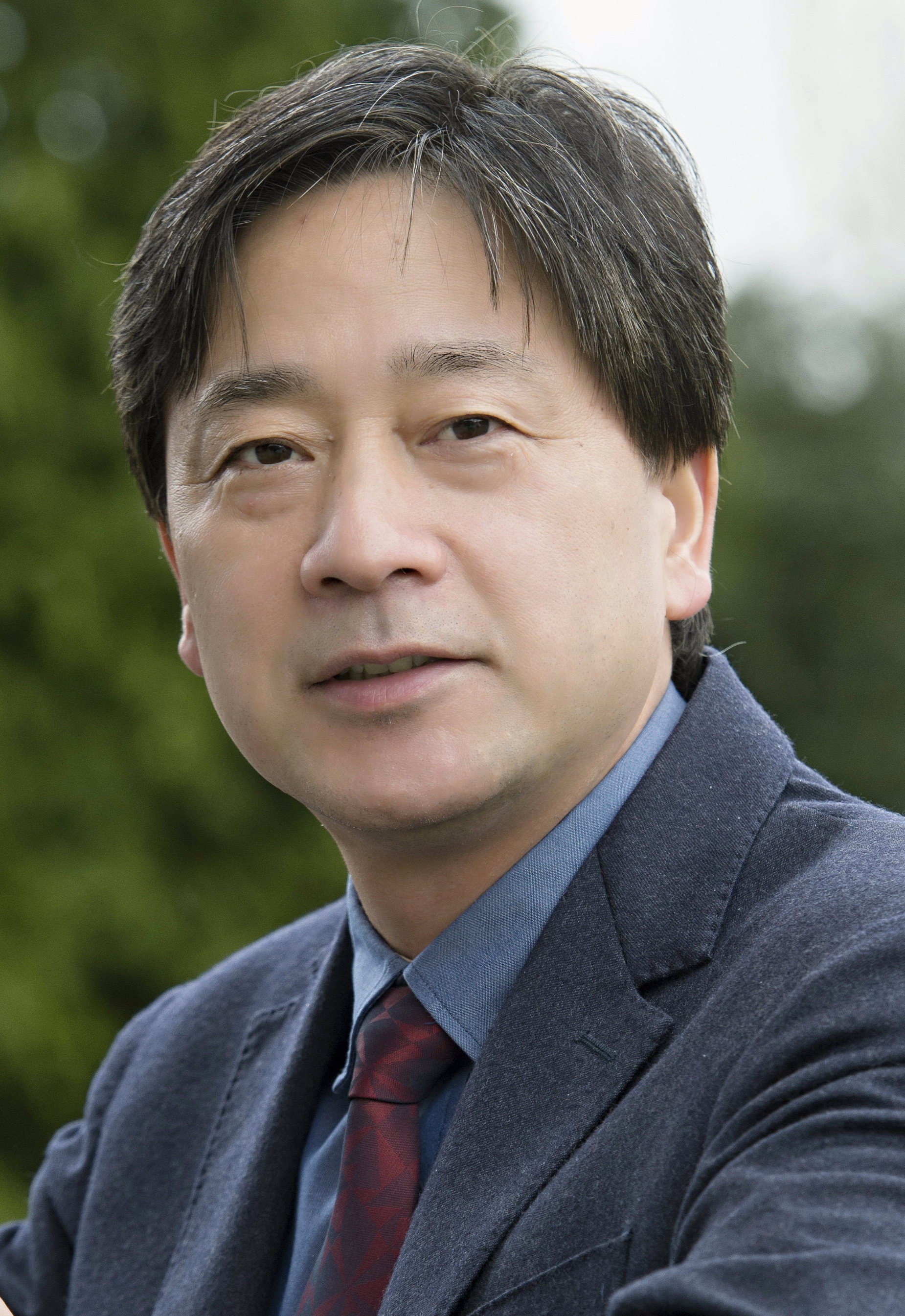}
\end{figure}

\noindent{\bf Yaochu Jin}\quad received the B.Sc., M.Sc., and Ph.D. degrees from Zhejiang University, Hangzhou, China, in 1988, 1991, and 1996, respectively, and the Dr.-Ing. degree from Ruhr University Bochum, Germany, in 2001. 
He is presently an Alexander von Humboldt Professor for Artificial Intelligence endowed by the German Federal Ministry of Education and Research, Chair of Nature Inspired Computing and Engineering, Faculty of Technology, Bielefeld University, Germany. He is also a Distinguished Chair, Professor in Computational Intelligence, Department of Computer Science, University of Surrey, Guildford, U.K. He was a “Finland Distinguished Professor” of University of Jyväskylä awarded by the Academy of Science and Finnish Funding Agency for Innovation, Finland, and “Changjiang Distinguished Visiting Professor” of Northeastern University, awarded by the Ministry of Education, China. His main research interests include human-centered learning and optimization, synergies between evolution and learning, and evolutionary developmental artificial intelligence. 
Prof. Jin is the President-Elect of the IEEE Computational Intelligence Society and the Editor-in-Chief of Complex \& Intelligent Systems. He was named by Clarivate as a “Highly Cited Researcher” from 2019 to 2022 consecutively. He is a Member of Academia Europaea and Fellow of IEEE.

E-mail: yaochu.jin@uni-bielefeld.de (Corresponding author)

ORCID iD: 0000-0003-1100-0631

\end{document}